\documentclass[runningheads]{llncs}

\usepackage{eccv}

\usepackage[dvipsnames]{xcolor}

\usepackage[utf8]{inputenc} 
\usepackage[T1]{fontenc}    
\usepackage{url}            
\usepackage{booktabs}       
\usepackage{amsfonts}       
\usepackage{nicefrac}       
\usepackage{microtype}      
\usepackage{xcolor}         

\usepackage{epsfig}
\usepackage{graphicx}
\usepackage{amsmath}
\usepackage{amssymb}

\usepackage{multirow}
\usepackage{bm}
\usepackage{float}
\usepackage{colortbl}
\usepackage{color, xcolor} 
\usepackage{comment}
\usepackage{algorithm}
\usepackage{algorithmicx}
\usepackage{algpseudocode}
\usepackage{indentfirst}
\usepackage{array}
\usepackage{booktabs}
\usepackage{soul}

\usepackage{arydshln}
\usepackage{wrapfig}
\newcommand{\PreserveBackslash}[1]{\let\temp=\\#1\let\\=\temp}
\newcolumntype{C}[1]{>{\PreserveBackslash\centering}p{#1}}
\newcolumntype{R}[1]{>{\PreserveBackslash\raggedleft}p{#1}}
\newcolumntype{L}[1]{>{\PreserveBackslash\raggedright}p{#1}}

\definecolor{remark}{rgb}{1,.5,0} 
\definecolor{citecolor}{rgb}{0,0.443,0.737} 
\definecolor{linkcolor}{rgb}{0.956,0.298,0.235} 
\definecolor{gray}{gray}{0.5}
\definecolor{cyan}{rgb}{0.831,0.901,0.945}
\definecolor{teal}{rgb}{0,0.5,0.5}
\definecolor{lightskyblue}{rgb}{0.53,0.8,0.976}
\definecolor{llgray}{rgb}{0.85,0.85,0.85}

\usepackage{pifont}
\usepackage{eccvabbrv}

\usepackage{graphicx}
\usepackage{booktabs}

\usepackage[accsupp]{axessibility}  

\usepackage{hyperref}

\usepackage{orcidlink}

\usepackage[capitalize]{cleveref}
\crefname{section}{Sec.}{Secs.}
\Crefname{section}{Section}{Sections}
\Crefname{table}{Table}{Tables}
\crefname{table}{Tab.}{Tabs.}

\colorlet{dark-blue}{blue!70!black}
\colorlet{dark-green}{green!60!black}
\colorlet{dark-red}{red!80!black}
\definecolor{mypink}{RGB}{219, 48, 122}
\hypersetup{
 colorlinks=true,%
 citecolor=citecolor,%
 filecolor=citecolor,%
 linkcolor=linkcolor,%
 urlcolor=magenta
}

\usepackage{lipsum}

\usepackage[normalem]{ulem}

\makeatletter\renewcommand\paragraph{\@startsection{paragraph}{4}{\z@}
  {.5em \@plus1ex \@minus.2ex}{-.5em}{\normalfont\normalsize\bfseries}}\makeatother

\newcolumntype{x}[1]{>{\centering\arraybackslash}p{#1pt}}
\newlength\savewidth

\usepackage{xspace}

\makeatletter
\DeclareRobustCommand\onedot{\futurelet\@let@token\@onedot}
\def\@onedot{\ifx\@let@token.\else.\null\fi\xspace}

\def\eg{\emph{e.g}\onedot} 
\def\ie{\emph{i.e}\onedot} 
\def\cf{\emph{c.f}\onedot} 
\def\etc{\emph{etc}\onedot}

\makeatother

\begin{document}

\title{UNIKD: UNcertainty-filtered Incremental Knowledge Distillation for Neural Implicit Representation}

\titlerunning{Uncertainty-filtered Incremental Knowledge Distillation for NIR}
\author{Mengqi Guo \and
Chen Li \and
Hanlin Chen \and
Gim Hee Lee
}

\authorrunning{Guo et al.}

\institute{Department of Computer Science, National University of Singapore
\\
\email{\{mengqi, gimhee.lee\}@comp.nus.edu.sg}
\\
\url{https://dreamguo.github.io/projects/UNIKD}
}

\maketitle

\renewcommand{\thefootnote}{\fnsymbol{footnote}}

\vspace{-3mm}
\begin{abstract}
    Recent neural implicit representations (NIRs) have achieved great success in the tasks of 3D reconstruction and novel view synthesis. However, they require the images of a scene from different camera views to be available for one-time training. This is expensive especially for scenarios with large-scale scenes and limited data storage. In view of this, we explore the task of incremental learning for NIRs in this work. We design a student-teacher framework to mitigate the catastrophic forgetting problem. Specifically, we iterate the process of using the student as the teacher at the end of each time step and let the teacher guide the training of the student in the next step. As a result, the student network is able to learn new information from the streaming data and retain old knowledge from the teacher network simultaneously. Although intuitive, naively applying the student-teacher pipeline does not work well in our task. Not all information from the teacher network is helpful since it is only trained with the old data. To alleviate this problem, we further introduce a random inquirer and an uncertainty-based filter to filter useful information. Our proposed method is general and thus can be adapted to different implicit representations such as neural radiance field (NeRF) and neural surface field. Extensive experimental results for both 3D reconstruction and novel view synthesis demonstrate the effectiveness of our approach compared to different baselines.
    \vspace{-1mm}
    \vspace{-1mm}
\end{abstract}

\vspace{-3mm}
\section{Introduction}
\vspace{-1mm}

Recent neural implicit representations (NIRs) \cite{mildenhall2020nerf, wang2021neus, yariv2021volume, yu2022monosdf} such as NeRF and neural surface field have attracted increasing attention in the last few years because of their great success in novel view synthesis and 3D reconstruction. The key to these representations is to memorize the volume density or SDF value
and view-dependent color of every spatial point in the scene with a multi-layer perceptron (MLP). Although the simple MLP networks implicitly represent the 3D scenes precisely, they require all images of a scene from different camera views to be available for a one-time training. This is expensive especially for scenarios with large-scale scenes and limited data storage.
In view of this limitation, we explore an important task of incremental learning for NIRs in this work. In the incremental setting, the model trains on the current data without accessing any previous data, but tests on both current and previous data.

\begin{figure*}[t]
    \centering
    \vspace{-1mm}
    \includegraphics[width=1.0\linewidth]{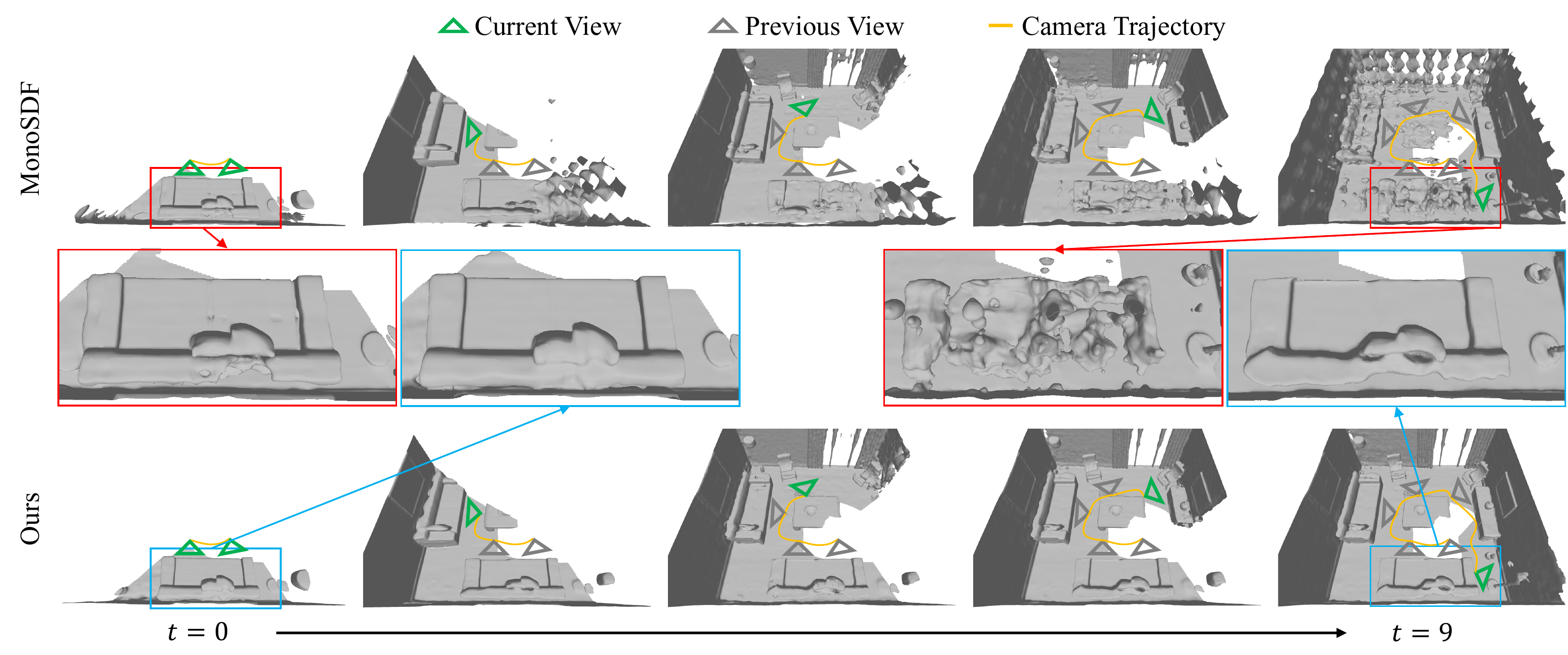}
    \vspace{-7mm}
    \caption{Visualization of the 3D reconstruction by MonoSDF \cite{yu2022monosdf} and our approach under the incremental setting.
    MonoSDF fails to reconstruct 3D surface observed at $t=0$ after being trained with new data because of the forgetting problem. In comparison, our approach is able to reconstruct both previously seen and new data.}
    \label{fig:teaser}
    \vspace{-6mm}
\end{figure*}

The main challenge for incremental learning is the catastrophic forgetting problem~\cite{robins1995catastrophic}, where the network trained on only new incoming images drastically forgets the previously learned knowledge. This is evident from the result of MonoSDF~\cite{yu2022monosdf} in Fig.~\ref{fig:teaser}, where the network is trained continuously with new incoming data captured along the yellow trajectory. The triangles represent camera views at different time steps, and green denotes the camera view at the current time step and gray ones for previous steps. We can see that the model fails to reconstruct the 3D scene observed at $t=0$ after being trained with new data. 
The catastrophic forgetting problem is widely discussed in the incremental learning literature \cite{aljundi2017expert, mallya2018packnet, aljundi2018memory, zhao2022static, li2017learning}, and the most related work to ours is Continual Neural Mapping (CNM) \cite{yan2021continual}. CNM is the first work that introduces the incremental setting for 3D reconstruction using the Signed Distance Function (SDF). A data-replay strategy~\cite{lee2020neural} is adopted in CNM to mitigate the forgetting problem. However, the data-replay still requires part of the previous training data to be stored.
Additionally, CNM only shows results for SDF-based 3D reconstruction, while we aim for a general pipeline for different NIRs such as NeRF and neural surface field.

In this work, we propose a student-teacher pipeline to tackle the catastrophic forgetting problem in incremental NIRs. Specifically, we first train the model with currently available data, and then use the trained model as the teacher model with knowledge distillation strategy~\cite{hinton2015distilling} to self-supervise the student network.
We iterate this process by using the student as the teacher at the end of each time step, and letting the teacher guide the training of the student in the next step. 
As a result, the student is able to learn from the newly available data and preserve the old knowledge from the teacher simultaneously. Furthermore, we also propose an alternate optimization strategy such that the new data and knowledge from the teacher can be effectively imparted to the student. 

The aim of introducing the teacher network is to impart the knowledge obtained from the previous training steps to the current student network. 
R2L~\cite{wang2022r2l} uses random input views to distill information from a well-trained NeRF to a compact network. 
However, the teacher network is trained only with the old views in our case and thus is unable to generate useful knowledge for the unseen views. 
To solve this problem, we further introduce a random inquirer and an uncertainty-based filter for filtering useful knowledge. 
We adopt the self-supervised uncertainty modeling from \cite{kendall2017uncertainties} to predict the uncertainty of the network for each input ray.
The inquirer randomly generates camera views for the uncertainty module and the filter removes the uncertain queries based on a confidence score. Intuitively, the uncertainty module would only have high confidence for the previously seen or similar data, and hence it is able to filter out the incorrect knowledge generated from the random query. 

We evaluate the effectiveness of our proposed approach on two popular NIRs, NeRF~\cite{mildenhall2020nerf} and MonoSDF~\cite{yu2022monosdf}. Extensive experimental results on both 3D reconstruction and novel view synthesis show that our approach mitigates the catastrophic forgetting problems effectively without storing previous training data. Specifically, our method significantly improves 39.6\% and 61.3\% over MonoSDF in terms of F1 on the large-scale datasets ICL-NUIM~\cite{handa2014ICL} and Replica~\cite{straub2019replica}. Moreover, our approach outperforms NeRF by 36.3\% and 63.9\% in terms of PSNR on the object-scale 360Capture~\cite{mildenhall2020nerf} and large-scale ScanNet~\cite{dai2017scannet} datasets, respectively.
Our contributions are summarized as follows:
\begin{itemize}
    \item We explore the incremental learning task for general NIRs.
    \item We propose a student-teacher pipeline to mitigate catastrophic forgetting in incremental learning.
    \item We design the uncertainty filter and the random inquirer to generate and select useful information for the student network.
    \item We significantly outperform baselines by a large margin for both 3D reconstruction and novel view synthesis. 
\end{itemize}

\section{Related Work}

\vspace{-1mm}
\paragraph{Neural Implicit Representation.}
The neural implicit representations (NIRs)~\cite{chen2023gnesf, li2021mine, wei2021nerfingmvs, yen2021inerf, lin2021barf} have shown remarkable potential in various computer vision tasks, such as novel view synthesis~\cite{mildenhall2020nerf, barron2023zip, barron2022mip} and 3D reconstruction~\cite{yu2022monosdf, azinovic2022neural, sun2021neuralrecon}. 
The pioneering work NeRF~\cite{mildenhall2020nerf} introduced a simple yet effective MLP network to implicitly capture the 3D scene and propose a differentiable rendering method for generating novel view images. 
Many follow-up works have attempted to enhance NeRF to fully exploit their potential, such as real-time rendering~\cite{yu2021plenoctrees, reiser2021kilonerf}, faster training~\cite{liu2020neural, fridovich2022plenoxels, muller2022instant, chen2022tensorf}, sparse view~\cite{yu2021pixelnerf, roessle2022dense}, generalizable model~\cite{wang2021ibrnet, chen2021mvsnerf}, lightning changing~\cite{martin2021nerf, mildenhall2022nerf}, better representation~\cite{barron2021mip, zhang2020nerf++}, \etc.
Some recent works~\cite{wang2021neus, yariv2021volume} proposed neural implicit surfaces and incorporated the signed distance function (SDF) into NeRF for smooth and accurate surface reconstruction. 
MonoSDF~\cite{yu2022monosdf} further leveraged monocular depth and normal priors to achieve more detailed reconstruction for larger 3D scenes. Despite the great success, existing NIRs suffer from the
catastrophic forgetting problem when continuously learning from streaming data. In view of this problem, we focus on the under-explored and yet important incremental settings for NIRs in this paper.

\vspace{-1mm}
\paragraph{Incremental Learning.}
Incremental learning is a classical machine learning problem where only partial data is available for training at each step. Existing methods typically fall into three categories \cite{de2021continual}: data replay~\cite{klein2007parallel, rebuffi2017icarl, rolnick2019experience, lopez2017gradient, chaudhry2018efficient, shin2017continual}, parameter regularization~\cite{li2017learning, rannen2017encoder, aljundi2018memory, kirkpatrick2017overcoming}, and parameter isolation~\cite{aljundi2017expert, xu2018reinforced, mallya2018packnet, fernando2017pathnet}. 
In this paper, we revisit some classical incremental approaches to build strong baselines. Specifically, PTAM~\cite{klein2007parallel} introduced keyframes replaying (KR) to avoid forgetting, MAS~\cite{aljundi2018memory} measured the parameter importance for each task and regularized the important parameters, PackNet~\cite{mallya2018packnet} assigned parameters subsets explicitly to different tasks by constituting binary masks, POD~\cite{douillard2020podnet} and AFC~\cite{kang2022class} employed knowledge distillation on the intermediate network features.
CNM~\cite{yan2021continual} is the first work on incremental learning for neural surface field, which reconstructs 3D surface from streaming depth inputs using a reply-based method~\cite{lee2020neural}.
CLNeRF~\cite{iccv23clnerf} applies the data replay on NeRF for novel view synthesis.
However, their method still requires access to some of the previous data to prevent forgetting and they mainly focus on the SDF or NeRF representation. In comparison, our approach can be adapted to different NIRs without access to any previous data. 

\vspace{-1mm}
\paragraph{NIR-SLAM and Large-scale NeRF.}
Traditional simultaneous localization and mapping
(SLAM)~\cite{newcombe2011kinectfusion, klein2009parallel, newcombe2011dtam} is able to reconstruct 3D scenes with streaming data, which share similar spirits with our incremental setting. Recently, some works such as iMAP~\cite{sucar2021imap} and NICE-SLAM~\cite{zhu2022nice} have adopted neural implicit representation as the scene representation in SLAM and achieved promising performance. These approaches mitigate the forgetting problem by storing keyframes, as done in traditional SLAM. The drawback of using keyframes is that memory usage will increase accordingly as the scene gets larger. On the other hand, Recent NeRFusion~\cite{zhang2022nerfusion} and Block-NeRF~\cite{tancik2022block} handle large-scale scenes by incrementally reconstructing a global scene
representation by fusing local voxel representations. However, the voxel-based representation requires substantial storage and the fusion stage requires all the images of a scene.  Moreover, all those approaches work on one specific neural implicit representation while we propose a general approach, which is also memory-efficient.

\section{Preliminaries}
\label{sec:Pre}

\vspace{-1mm}
\subsection{Neural Implicit Representations}
In this section, we present a unified formulation for the currently dominant NIRs including NeRF~\cite{mildenhall2020nerf} and neural surface field~\cite{yariv2021volume, yu2022monosdf}. The principal idea is to use a simple neural network such as MLPs to memorize the color $\textbf{c}=(r,g,b)$, volume density $\sigma$ for each location $\textbf{x}=(x,y,z)$ and camera view direction $\textbf{d}=(\theta, \phi)$ in a 3D scene. 
While existing neural surface field predicts SDF value which is then converted to density, we use $(\textbf{c},\sigma)=F(\textbf{x}, \textbf{d})$ to represent both NeRF and neural surface field networks in this paper for simplicity.
The per-pixel RGB $c(\textbf{r})$ value of an image can be rendered with $N$ 3D points taken along the ray $\textbf{r}$ from the camera center to the pixel as:
\begin{equation}
c(\textbf{r}) = \sum_{i=1}^{N}T_{i}\big(1-\exp(-\sigma_{i}\delta_{i})\big)\textbf{c}_{i},
\label{eq:rendering}
\end{equation}
where $T_{i}=\exp(-\sum_{j=1}^{i-1}\sigma_{j}\delta_{j})$ is the accumulated transmittance along the ray $\textbf{r}$ from camera center to $i^{\text{th}}$ 3D point, $\delta_{i}$ indicates the distance between $i^{\text{th}}$ sample and $(i+1)^{\text{th}}$ sample. Since the whole pipeline is differentiable, the rendering output can be directly supervised by the RGB image:
\begin{equation}
\label{eq:pre_opt}
\begin{aligned}
\mathcal{L} = \mathcal{L}_{rgb} + [\mathcal{L}_{eik}] + [\mathcal{L}_{prior}],
\quad \text{where} \quad
\mathcal{L}_{rgb} = \sum_{\textbf{r}\in R} \Bigl(||c^*(\textbf{r})-c(\textbf{r})||^2_2\Bigr)
\end{aligned}
\end{equation}
represents the rendering loss. $c^*(\textbf{r})$ denotes the ground truth color and $R$ represents a group of rays from one or more camera views.
$\mathcal{L}_{eik}$ is the SDF regularizer for neural surface field~\cite{wang2021neus, yariv2021volume, yu2022monosdf}, and $\mathcal{L}_{prior}$ represents the geometry prior term~\cite{yu2022monosdf}. 
Generally, the prior term consists of depth and normal priors. More details about the depth and normal priors are provided in supplementary material.
Note that $\mathcal{L}_{rgb}$ is the fundamental term for both NeRF and neural surface field. $\mathcal{L}_{eik}$ and $\mathcal{L}_{prior}$ are prior terms used in \cite{yu2022monosdf}, which we denote with $[.]$. 

\vspace{-1mm}
\subsection{Catastrophic Forgetting of NIRs}
Despite the impressive performance, existing NIRs require training on the entire set of images covering all views. In practice, this may not be feasible in the scenario of limited data storage or streaming data, which may require the network to be trained on new data without revisiting old ones. This may lead to the catastrophic forgetting of existing NIRs, where the network quickly forgets previously learned knowledge while acquiring new knowledge. To address this issue, we explore the task of incremental learning for NIRs with the goal of mitigating the catastrophic forgetting problem.

\vspace{-1mm}
\section{Our Method}
\vspace{-1mm}
In this section, we first introduce the incremental setting for NIRs. We then represent our proposed student-teacher pipeline with an uncertainty based filter and an alternative optimization strategy.

\begin{figure*}[t]
\centering
\vspace{-1mm}
\includegraphics[width=1.0\linewidth]{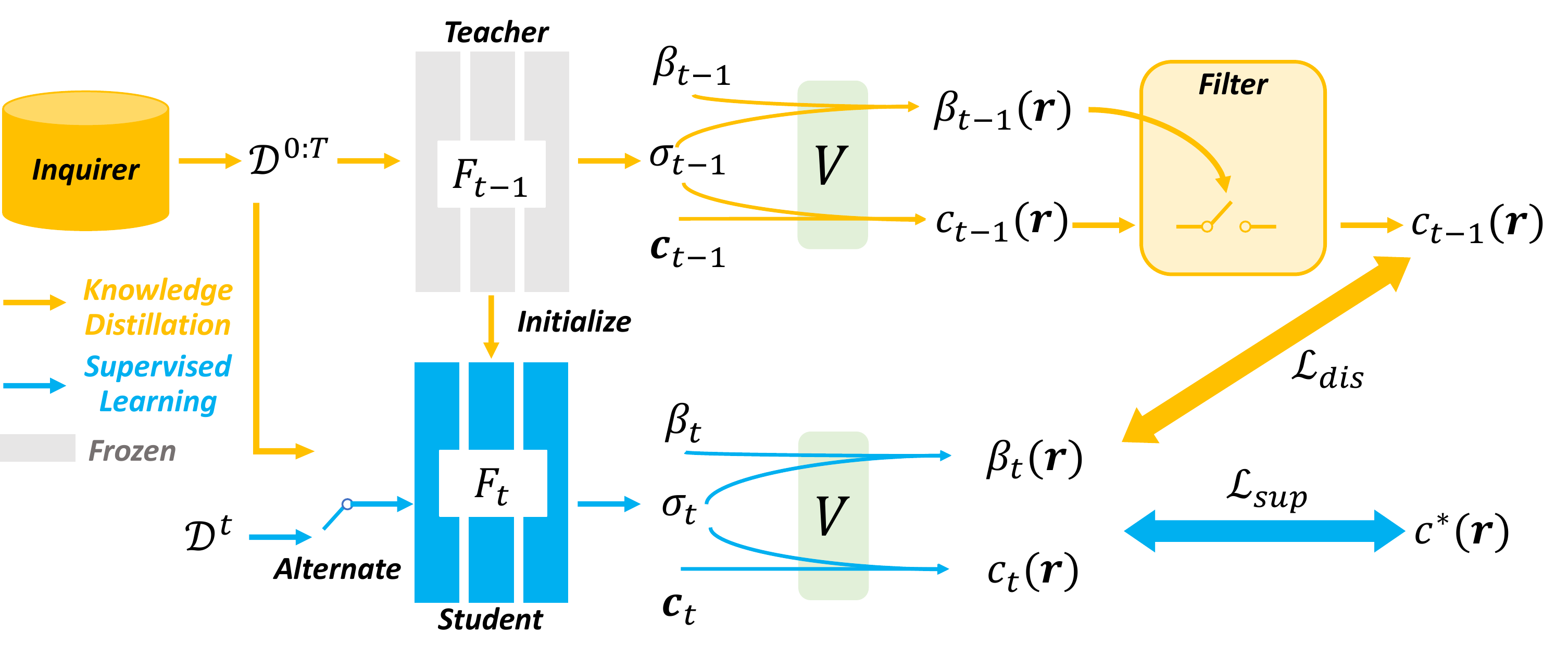}
    \vspace{-7mm}
    \caption{The overall framework of our proposed student-teacher pipeline. At time step $t$, The student network learns simultaneously from the currently available data $\mathcal{D}^t$ and the previously learned knowledge from the teacher network. The input of the teacher network is generated with the random inquirer. The output is filtered with an uncertainty based filter for useful information selection. $V$ denotes the differentiable volume renderer.}
    \label{fig:framework}
    \vspace{-6mm}
\end{figure*}

\vspace{-1mm}
\subsection{Problem Definition}
\vspace{-1mm}
We consider a common scenario in the robotics or vision community, where $T+1$ groups of data $\mathcal{D}=\{\mathcal{D}^0,\mathcal{D}^1,\cdots,\mathcal{D}^T\}$ come in sequentially. The data for each time step $t$ consists of $N$ pairs of images $I^t$ and the corresponding camera poses $P^t$, \ie, $\mathcal{D}^t=\{d^t_0,d^t_1,\cdots,d^t_N\}$, where $d^t_n=(I^t_n,P^t_n)$. Generally, the camera poses do not overlap between different time steps.
The task of incremental learning for NIRs aims to continually learn from the newly arriving data $\mathcal{D}^{t}$ and also preserve the knowledge from previously seen data $\mathcal{D}^{0:t-1}$.

\vspace{-1mm}
\subsection{Overview}
\vspace{-1mm}
The overall framework of our approach is illustrated in Fig.~\ref{fig:framework}. It comprises a student and a teacher network, both sharing the same architecture with a density branch, a color branch, and an uncertainty branch.
At each time step $t$, the student learns simultaneously from the currently available data $\mathcal{D}^{t}$ and the previously learned knowledge from the teacher network. The student model trained in this step is then utilized as the teacher model in the next step and imparts its acquired knowledge to the next student. We iterate this process throughout the training process. To explore the knowledge space of the teacher network, we design a random inquirer that generates camera views for the teacher network. However, the teacher network can generate erroneous information for randomly generated views because it only trains on previously seen data $\mathcal{D}^{0:t-1}$. We further design an uncertainty branch to predict uncertainty scores and select only reliable information from the teacher network. 

\vspace{-1mm}
\subsection{Supervised Learning}
\vspace{-1mm}
At each time step $t$, we utilize the available data $\mathcal{D}^{t}$ to train the student network directly with rendering loss. Except for the density and color, we also predict the uncertainty value for each input to measure the confidence.

\vspace{-1mm}
\paragraph{Uncertainty Modeling.}
\label{Sec:uncertainty}
We adopt the self-supervised uncertainty formulation from \cite{kendall2017uncertainties} to model the uncertainty of the network for each input ray. This formulation has also been used in previous NeRF-W~\cite{martin2021nerf} to distinguish the static and transient scenes. With a different objective, we aim to indicate the confidence of the network on the current input.
Specifically, we build an additional branch that shares the same input as the color branch to predict the uncertainty $(\mathbf{c},\sigma,\beta)=F(\textbf{x},\textbf{d})$. We adopt Softplus as the activation function on the uncertainty for stable training.
Finally, we compute the pixel-wise uncertainty from each sample point using the same volume rendering technique as the color:
\begin{equation}
\begin{aligned}
\beta(\textbf{r}) = \sum_{n=1}^{N}T_{i}\big(1-exp(-\sigma_{j}\delta_{j})\big)\hat{\beta}_{i} + \beta_{min},
\quad \text{where} \quad
\hat{\beta}_i = \log\big(1+e^{\beta_i-1}\big), 
\end{aligned}
\label{eq:renderingBeta}
\end{equation}
$T_i$ represents the accumulated transmittance expressed by Eqn.~\eqref{eq:rendering}, and $\beta_{min}$ denotes a hyper-parameter that ensures the minimum uncertainty following \cite{martin2021nerf}.

\vspace{-1mm}
\paragraph{Supervised Optimization.}
The objective function of supervised training is:
\begin{equation}
\label{eq:sup}
\begin{aligned}
\mathcal{L}_{sup} =
\sum_{\textbf{r}\in R} \Big(
\frac{||c^*(\textbf{r})-c_{t}(\textbf{r})||^2_2}{2} + 
\frac{||c^*(\textbf{r})-c_{t}(\textbf{r})||^2_2}{2*\beta_t(\textbf{r})^2} 
+ \log\big(\beta_t(\textbf{r})\big) + \eta \Big),
\end{aligned}
\end{equation}
where $\eta$ denotes the margin of the uncertainty regular term to avoid negative values. Note that we only need to supervise the color and the uncertainty is implicitly learned from the loss function. Intuitively, on one hand, the network needs to predict a high uncertainty value when the color prediction is inaccurate in order to minimize the loss function. On the other hand, the regularization term $\log(\beta_t(\textbf{r}))$ prevents the network from predicting infinite uncertainty. 

\subsection{Knowledge Distillation}
The network trained with only the currently available data at each time step tends to forget the previously learned knowledge, referred as the catastrophic forgetting
problem. To prevent this, we further introduce a teacher network to impart the previously learned knowledge to the current model. 

\paragraph{Student-teacher Modeling.}
\label{Sec:Student-teacher}
At each time step $t$, the student network $F_{t}$ concurrently learns from the teacher network $F_{t-1}$ and the new coming data $\mathcal{D}^{t}$. The student network is then used as the teacher network after each step. We iterate the process of using the student as the teacher at the end of each step, and let the teacher guide the student in the next step.
As a result, the student network can learn both new knowledge from $\mathcal{D}^{t}$ and old knowledge from $F_{t-1}$ and hence mitigate the forgetting problem.
To facilitate the knowledge imparting from the teacher to the student network, we introduce a knowledge distillation loss~\cite{hinton2015distilling}. Moreover, we initialize the parameters of the student network with that of the teacher network. This initialization strategy also helps to mitigate the forgetting problem since the parameters are learned from previously seen data.

\paragraph{Random Inquirer.}
The role of the teacher network is to impart old knowledge, which means the inputs should be the same as the training data from the previous time steps. However, the previous training data is not accessible under the incremental setting. To solve this problem, we design a constrained random inquirer to generate inputs for the teacher network. 
For 
scenes where the camera moves along a trajectory, we randomly generate camera views in the range of each degree of freedom of the camera matrix from the previous data. Specifically, we store the range of six values in the camera matrix, \ie, $r=(x_{min}, x_{max}, y_{min}, y_{max}, z_{min}, z_{max}, \alpha_{min}, \alpha_{max}, \beta_{min}, \beta_{max}, \gamma_{min}, \gamma_{max})$, at each time step $t$. We first randomly choose a time step $k$ from $0:t-1$ at time step $t$, and then randomly generate a group of the six values in the range of $r_k$ to compute the camera matrix.

\paragraph{Uncertainty-based Filter.}
\label{Sec:uncertainty-filter}
The role of the random inquirer is to explore the knowledge space of the teacher network such that we can extract useful information, \ie the knowledge from previous time steps. However, the teacher network might output incorrect knowledge since it has been trained only on $D^{0:t-1}$, while the input generated from the random inquirer covers the whole dataset. To overcome this issue, we utilize the uncertainty module as described in Sec.~\ref{Sec:uncertainty} for useful knowledge selection. Specifically, we take the average of the output uncertainty value over rays from one camera view, and only select camera views with an uncertainty smaller than a threshold $\beta^{thr}$, \ie:
\begin{equation}
\label{eq:beta_thr}
\begin{aligned}
R^* \leftarrow R^* \cup R_v, \quad \text{if} \quad \frac{1}{\mathcal{N}_{R_v}}\sum_{\textbf{r}\in R_v}\big(\beta_{t}(\textbf{r})\big) < \beta^{thr}.
\end{aligned}
\end{equation}
$R_v$ is the rays for camera view $v$ generated from the random inquirer, $\mathcal{N}_{R_v}$ is the number of rays samples, and $R^*$ is the collection of data samples we use for knowledge distillation. Intuitively, the network tends to output lower uncertainty for previously seen data compared to unseen ones, and thus we can use $R^*$ to approximate the unavailable data from the previous training step.
Note that the selection is conducted in terms of camera views, \ie average over all ray samples instead of a single ray. This is empirically shown to better distinguish the seen and unseen images.

\vspace{1mm}
\paragraph{Distilled Optimization.}
Finally, we use the teacher network to guide the student network via a knowledge distillation loss:
\begin{equation}
\label{eq:dis}
\begin{aligned}
\mathcal{L}_{dis} =\sum_{\textbf{r}\in R^*} \Bigl(
\frac{||c_{t-1}(\textbf{r})-c_{t}(\textbf{r})||^2_2}{2} + 
\frac{||c_{t-1}(\textbf{r})-c_{t}(\textbf{r})||^2_2}{2*\beta_{t}(\textbf{r})^2} 
+ \log\big(\beta_{t}(\textbf{r})\big) + \eta \Bigr),
\end{aligned}
\end{equation}
where $c_{t-1}(\textbf{r})$ and $c_{t}(\textbf{r})$ represent the output color of the teacher and student model, respectively. $R^*$ denotes useful data selection from Eqn.~\eqref{eq:beta_thr}. With knowledge distillation, the student network is able to preserve the previously learned knowledge throughout the whole training process.

\vspace{1mm}
\subsection{Iterative Optimization}
We propose an iterative optimization mechanism to enable the student network to learn simultaneously from the current data $\mathcal{D}^{t}$ and knowledge from the teacher network. Specifically, we alternatively optimize the supervised loss Eqn.~\eqref{eq:sup} and the knowledge distillation loss Eqn.~\eqref{eq:dis}, \ie:
\begin{equation}
\label{eq:loss}
\begin{aligned}
\mathcal{L} = 
    \begin{cases}
    	\mathcal{L}_{sup} + [\mathcal{L}_{eik}] + [\mathcal{L}_{prior}], &\text{if}\ i\ \text{is even} \\
    	\mathcal{L}_{dis} + [\mathcal{L}_{eik}] + [\mathcal{L}_{prior\_dis}], &\text{otherwise}
    \end{cases}
,
\end{aligned}
\end{equation}
where $i$ denotes the iteration number, $\mathcal{L}_{prior\_dis}$ represents that the prior comes from the teacher network instead of the pre-trained model as in the $\mathcal{L}_{prior}$.

\vspace{-2mm}
\section{Experiments}

\renewcommand\arraystretch{1.2}
\begin{table*}[t]
\centering
\caption{Comparison with baselines on the ICL-NUIM. (Best and second best results are highlighted in bold and underlined, respectively.)
}
\vspace{-3mm}
\resizebox{0.95\linewidth}{!}{\begin{tabular}{c|cc|ccccccc}
\hline
& MonoSDF~\cite{yu2022monosdf} & MonoSDF*~\cite{yu2022monosdf} & CNM~\cite{yan2021continual} & MAS~\cite{aljundi2018memory} & PackNet~\cite{mallya2018packnet} & KR~\cite{klein2007parallel}  & POD~\cite{douillard2020podnet}  & AFC~\cite{kang2022class} & Ours  \\
\hline
F1$\uparrow$ & ${64.71}$ & ${89.68}$ & ${69.93}$ & ${66.40}$ & ${76.03}$ & $\underline{86.78}$ & ${84.52}$ & ${86.17}$ & $\textbf{90.32}$\\
CD$\downarrow$ & ${5.94}$ & ${2.64}$ & ${5.42}$ & ${5.84}$ & ${4.39}$ & $\underline{3.02}$ & ${3.32}$ & ${3.10}$ & $\textbf{2.60}$ \\
\hline
\end{tabular}}
\small
\label{Tab:exp_ICL}
\vspace{-3mm}
\end{table*}

\begin{table*}[t]
\centering
\small
\setlength{\tabcolsep}{0.08cm}
\caption{Comparison with baselines on the Replica. (Best and second best results are highlighted in bold and underlined, respectively.)
}
\vspace{-3mm}
\resizebox{\linewidth}{!}{\begin{tabular}{c|cc|cc:ccccccc}
\hline
& MonoSDF~\cite{yu2022monosdf} & MonoSDF*~\cite{yu2022monosdf} & iMAP~\cite{sucar2021imap} & NICE-SLAM~\cite{zhu2022nice} & CNM~\cite{yan2021continual} & MAS~\cite{aljundi2018memory} & PackNet~\cite{mallya2018packnet} & KR~\cite{klein2007parallel} & POD~\cite{douillard2020podnet} & AFC~\cite{kang2022class} & Ours \\
\hline
F1$\uparrow$ & ${53.63}$ & ${86.18}$ & - & - & ${67.52}$ & ${58.75}$ & ${61.99}$ & $\underline{79.67}$ & ${72.59}$ & ${74.96}$ & $\textbf{86.52}$ \\
CD$\downarrow$ & ${8.58}$ & ${2.94}$ & ${4.99}$ & $\textbf{2.93}$ & ${6.54}$ & ${7.98}$ & ${7.49}$ & ${3.99}$ & ${4.56}$ & ${4.20}$ & $\underline{3.11}$ \\
\hline
\end{tabular}}
\label{Tab:exp_replica}
\vspace{-4mm}
\end{table*}

\vspace{-1mm}
\subsection{Experimental Settings}
\vspace{-1mm}
We apply our approach to currently dominant implicit representations NeRF~\cite{mildenhall2020nerf} and neural surface field ~\cite{yu2022monosdf}, and show results for both novel view synthesis and 3D reconstruction.

\vspace{-1mm}
\paragraph{Dataset.}
The previous NIRs conducted experiments on different types of scenes, thus we consider the following datasets to cover: 
a) Object-scale scenes, \ie 360Capture~\cite{mildenhall2020nerf};
b) Large-scale synthetic scenes, \ie ICL-NUIM~\cite{handa2014ICL} and Replica~\cite{straub2019replica};
c) Large-scale real-world scenes, \ie ScanNet~\cite{dai2017scannet}.
For incremental setting, we divide the images of each scene and the corresponding camera poses into 10 time steps $\mathcal{D}=\{\mathcal{D}^0,\mathcal{D}^1,\cdots,\mathcal{D}^9\}$.

\vspace{-1mm}
\paragraph{Baselines.}
We compare against 
a) the main baselines NeRF~\cite{mildenhall2020nerf} and MonoSDF~\cite{yu2022monosdf} 
under both incremental and batch training settings;
b) SLAM-based NIRs iMAP~\cite{sucar2021imap} and NICE-SLAM~\cite{zhu2022nice};
c) Reply-based NIRs CNM~\cite{yan2021continual} and KR~\cite{klein2007parallel} (replay 10 keyframes following iMAP~\cite{sucar2021imap}), ;
d) Four representative incremental learning baselines MAS~\cite{aljundi2018memory}, PackNet~\cite{mallya2018packnet}, POD~\cite{douillard2020podnet}, AFC~\cite{kang2022class}. 

\vspace{-1mm}
\paragraph{Evaluation Metrics.}
For 3D reconstruction, we follow MonoSDF~\cite{yu2022monosdf} to report Chamfer Distance (CD) and F1 score with a threshold of 5cm.
For novel view synthesis, we follow NeRF~\cite{mildenhall2020nerf} to report PSNR, SSIM, and LPIPS.

\vspace{-1mm}
\paragraph{Backbone for 3D Reconstruction.}
In the 3D reconstruction experiments, we adopt MonoSDF~\cite{yu2022monosdf} as our backbone. Our network $F$ consists of an SDF network, a color network, and an uncertainty network. The density network consists of eight fully connected (FC) layers with $256$-channel, the color and uncertainty networks are both two FC layers with $256$-channel. We train our network with a batch size of $1024$ rays, where each ray samples $96$ points.

\vspace{-1mm}
\paragraph{Backbone for Novel View Synthesis.}
In the novel view synthesis experiments, we adopt NeRF~\cite{mildenhall2020nerf} as our backbone. Specifically, our network $F$ consists of a density network, a color network, and an uncertainty network. The density network consists of eight fully connected (FC) layers with $256$-channel, the color and uncertainty networks are both one FC layer with $128$-channel. We train our network with a batch size of $1024$ rays, $64$ points per ray for the coarse network and $64+128$ points per ray for the fine network. 
We adopt AlexNet~\cite{krizhevsky2017imagenet} to compute LPIPS.

\begin{table*}[t]
\centering
\small
\vspace{-1mm}
\setlength{\tabcolsep}{0.08cm}
\caption{Quantitative comparison with baselines on the 360Capture dataset. We show results for each step test datasets $\mathcal{D}^0, \mathcal{D}^3, \mathcal{D}^6, \mathcal{D}^9$ and the average performance over ${\mathcal{D}^{0:9}}$, respectively. All models are incrementally trained on the 10-step training datasets.
}
\vspace{-3mm}
\resizebox{\linewidth}{!}{\begin{tabular}{c|cccc|c}
\hline
\multirow{2}{*}{Method} &\multicolumn{5}{c}{Test Dataset (PSNR$\uparrow$ / SSIM$\uparrow$ / LPIPS$\downarrow$)} \\
\cline{2-6}
& ${\mathcal{D}^0}$ & ${\mathcal{D}^3}$ & ${\mathcal{D}^6}$ & ${\mathcal{D}^9}$ & Average on ${\mathcal{D}^{0:9}}$ \\
\hline
NeRF~\cite{mildenhall2020nerf} & 
${16.10}$ / ${0.468}$ / ${0.276}$ & 
${14.97}$ / ${0.407}$ / ${0.357}$ & 
${15.50}$ / ${0.512}$ / ${0.323}$ & 
${18.18}$ / ${0.636}$ / ${0.217}$ & 
${15.56}$ / ${0.468}$ / ${0.317}$  \\
NeRF*~\cite{mildenhall2020nerf} & 
${24.27}$ / ${0.781}$ / ${0.177}$ & 
${23.81}$ / ${0.767}$ / ${0.184}$ &  
${22.86}$ / ${0.738}$ / ${0.188}$ & 
${20.75}$ / ${0.684}$ / ${0.227}$ & 
${22.81}$ / ${0.741}$ / ${0.198}$  \\
\hline
MAS~\cite{aljundi2018memory} & 
${18.15}$ / ${0.552}$ / ${0.273}$ & 
${16.42}$ / ${0.500}$ / ${0.332}$ &  
${16.95}$ / ${0.525}$ / ${0.374}$ & 
${17.35}$ / ${0.551}$ / ${0.340}$ & 
${17.02}$ / ${0.513}$ / ${0.341}$  \\
PackNet~\cite{mallya2018packnet} & 
${17.21}$ / ${0.497}$ / ${0.349}$ & 
${17.09}$ / ${0.491}$ / ${0.378}$ &  
${17.01}$ / ${0.489}$ / ${0.390}$ & 
${14.45}$ / ${0.417}$ / ${0.447}$ & 
${16.52}$ / ${0.474}$ / ${0.388}$  \\
KR~\cite{klein2007parallel} & 
${18.76}$ / ${0.629}$ / ${0.225}$ & 
${19.50}$ / ${0.625}$ / ${0.238}$ & 
$\textbf{21.14}$ / $\textbf{0.682}$ / ${0.213}$ & 
${19.46}$ / ${0.657}$ / ${0.328}$ &  
${19.68}$ / ${0.642}$ / ${0.240}$  \\
POD~\cite{douillard2020podnet} & 
${18.54}$ / ${0.585}$ / ${0.269}$ & 
${17.29}$ / ${0.502}$ / ${0.317}$ &  
${18.17}$ / ${0.542}$ / ${0.270}$ & 
${18.31}$ / ${0.573}$ / ${0.291}$ & 
${17.79}$ / ${0.561}$ / ${0.291}$  \\
AFC~\cite{kang2022class} & 
${19.01}$ / ${0.603}$ / ${0.258}$ & 
${19.19}$ / ${0.621}$ / ${0.252}$ &  
${19.37}$ / ${0.639}$ / ${0.240}$ & 
${19.45}$ / ${0.643}$ / ${0.233}$ & 
${19.30}$ / ${0.632}$ / ${0.241}$  \\
Ours & 
$\textbf{22.48}$ / $\textbf{0.701}$ / $\textbf{0.188}$ & 
$\textbf{22.16}$ / $\textbf{0.694}$ / $\textbf{0.208}$ & 
${21.07}$ / $\textbf{0.682}$ / $\textbf{0.173}$ & 
$\textbf{19.82}$ / $\textbf{0.658}$ / $\textbf{0.221}$ & 
$\textbf{21.21}$ / $\textbf{0.672}$ / $\textbf{0.211}$ \\
\hline
\end{tabular}}
\label{Tab:exp_real}
\end{table*}

\begin{table*}[t]
\centering
\small
\vspace{-2mm}
\setlength{\tabcolsep}{0.08cm}
\caption{Quantitative comparison with baselines on the ScanNet dataset. We show results for each step test datasets $\mathcal{D}^0, \mathcal{D}^3, \mathcal{D}^6, \mathcal{D}^9$ and the average performance over $\mathcal{D}^{0:9}$, respectively. All models are incrementally trained on the 10-step training datasets.
}
\vspace{-3mm}
\resizebox{\linewidth}{!}{\begin{tabular}{c|cccc|c}
\hline
\multirow{2}{*}{Method} &\multicolumn{5}{c}{Test Dataset (PSNR$\uparrow$ / SSIM$\uparrow$ / LPIPS$\downarrow$)} \\
\cline{2-6}
& ${\mathcal{D}^0}$ & ${\mathcal{D}^3}$ & ${\mathcal{D}^6}$ & ${\mathcal{D}^9}$ & Average on ${\mathcal{D}^{0:9}}$ \\
\hline
NeRF~\cite{mildenhall2020nerf} & 
${12.61}$ / ${0.580}$ / ${0.396}$ & 
${11.59}$ / ${0.505}$ / ${0.571}$ & 
${15.65}$ / ${0.578}$ / ${0.470}$ & 
${25.95}$ / ${0.876}$ / ${0.145}$ & 
${13.78}$ / ${0.576}$ / ${0.460}$  \\
NeRF*~\cite{mildenhall2020nerf} & 
${22.55}$ / ${0.824}$ / ${0.244}$ & 
${22.23}$ / ${0.850}$ / ${0.231}$ &  
${24.53}$ / ${0.860}$ / ${0.211}$ & 
${25.51}$ / ${0.881}$ / ${0.150}$ & 
${23.55}$ / ${0.852}$ / ${0.212}$  \\
\hline
iMAP~\cite{sucar2021imap} & 
${10.85}$ / ${0.581}$ / ${0.472}$ & 
${19.53}$ / ${0.808}$ / ${0.300}$ & 
${19.19}$ / ${0.789}$ / ${0.326}$ & 
${20.88}$ / ${0.794}$ / ${0.328}$ &  
${18.80}$ / ${0.761}$ / ${0.340}$  \\
\hdashline
MAS~\cite{aljundi2018memory} & 
${16.40}$ / ${0.669}$ / ${0.366}$ & 
${12.15}$ / ${0.595}$ / ${0.512}$ &  
${15.76}$ / ${0.684}$ / ${0.455}$ & 
${22.22}$ / ${0.812}$ / ${0.303}$ & 
${15.76}$ / ${0.673}$ / ${0.402}$  \\
PackNet~\cite{mallya2018packnet} & 
${12.74}$ / ${0.535}$ / ${0.488}$ & 
${11.99}$ / ${0.604}$ / ${0.456}$ &  
${13.60}$ / ${0.630}$ / ${0.429}$ & 
${11.49}$ / ${0.550}$ / ${0.457}$ & 
${12.75}$ / ${0.592}$ / ${0.439}$  \\
KR~\cite{klein2007parallel} & 
${14.03}$ / ${0.598}$ / ${0.392}$ & 
${12.72}$ / ${0.583}$ / ${0.525}$ &  
${16.69}$ / ${0.691}$ / ${0.396}$ & 
${22.02}$ / ${0.801}$ / ${0.271}$ & 
${16.04}$ / ${0.663}$ / ${0.397}$  \\
POD~\cite{douillard2020podnet} & 
${15.23}$ / ${0.639}$ / ${0.399}$ & 
${12.83}$ / ${0.597}$ / ${0.472}$ &  
${15.27}$ / ${0.660}$ / ${0.420}$ & 
${22.10}$ / ${0.812}$ / ${0.268}$ & 
${15.35}$ / ${0.661}$ / ${0.407}$  \\
AFC~\cite{kang2022class} & 
${16.28}$ / ${0.657}$ / ${0.412}$ & 
${14.92}$ / ${0.640}$ / ${0.437}$ &  
${16.32}$ / ${0.663}$ / ${0.408}$ & 
${22.58}$ / ${0.843}$ / ${0.254}$ & 
${16.38}$ / ${0.684}$ / ${0.382}$  \\
Ours & 
$\textbf{21.74}$ / $\textbf{0.812}$ / $\textbf{0.224}$ & 
$\textbf{20.21}$ / $\textbf{0.825}$ / $\textbf{0.296}$ & 
$\textbf{23.90}$ / $\textbf{0.851}$ / $\textbf{0.224}$ & 
$\textbf{25.30}$ / $\textbf{0.876}$ / $\textbf{0.162}$ & 
$\textbf{22.59}$ / $\textbf{0.841}$ / $\textbf{0.230}$ \\
\hline
\end{tabular}}
\vspace{-2mm}
\label{Tab:exp_scannet}
\end{table*}

\vspace{-1mm}
\subsection{Results on 3D Reconstruction}
We first evaluate our approach for the neural surface field based representation. 
We adopt MonoSDF as the backbone and show 3D reconstruction results on the large-scale datasets ICL-NUIM and Replica. 

\vspace{-1mm}
\paragraph{ICL-NUIM.}
We show the results of our approach and the baselines on the ICL-NUIM dataset in Tab.~\ref{Tab:exp_ICL}. We can see that the performance of MonoSDF drops significantly when trained under incremental setting compared to the results under batch training (MonoSDF*). Note that batch training means that all data are available for one-time training, which is the upper bound of incremental training. The incremental baselines only achieve minor improvement compared with MonoSDF with the exception of KR. However, KR requires more memory as shown in Tab.~\ref{Tab:memory}. In comparison, our approach improves over the MonoSDF baseline by 39.6\% in F1 and is even slightly better than batch training while keeping a low memory usage.

\vspace{-1mm}
\paragraph{Replica.}
We further show results on the Replica dataset in Tab.~\ref{Tab:exp_replica}. We can see that MonoSDF and incremental baselines (MAS, PackNet, and KR) suffer from the forgetting problem, which can be evident from the performance drop compared to MonoSDF*. Our approach improves MonoSDF by 61.3\% for F1 and also achieves similar performance with batch training.
Compared to the SLAM-based baselines, our model outperforms iMAP by a large margin and is only slightly worse than NICE-SLAM. The better performance of NICE-SLAM can be attributed to the use of more powerful representations compared to our backbone MonoSDF. 
Moreover, the memory usage of SLAM-based baselines is larger than ours and increases as the scene gets larger, as discussed in Sec.~\ref{sec:memory}. 
Note that we compare with different baselines on the two datasets since CNM and SLAM-based NIRs (iMAP, NICE-SLAM) show results on the ICL-NUIM and Replica, respectively.

\begin{figure*}[t]
\centering
\vspace{-2mm}
\includegraphics[width=1.0\linewidth]{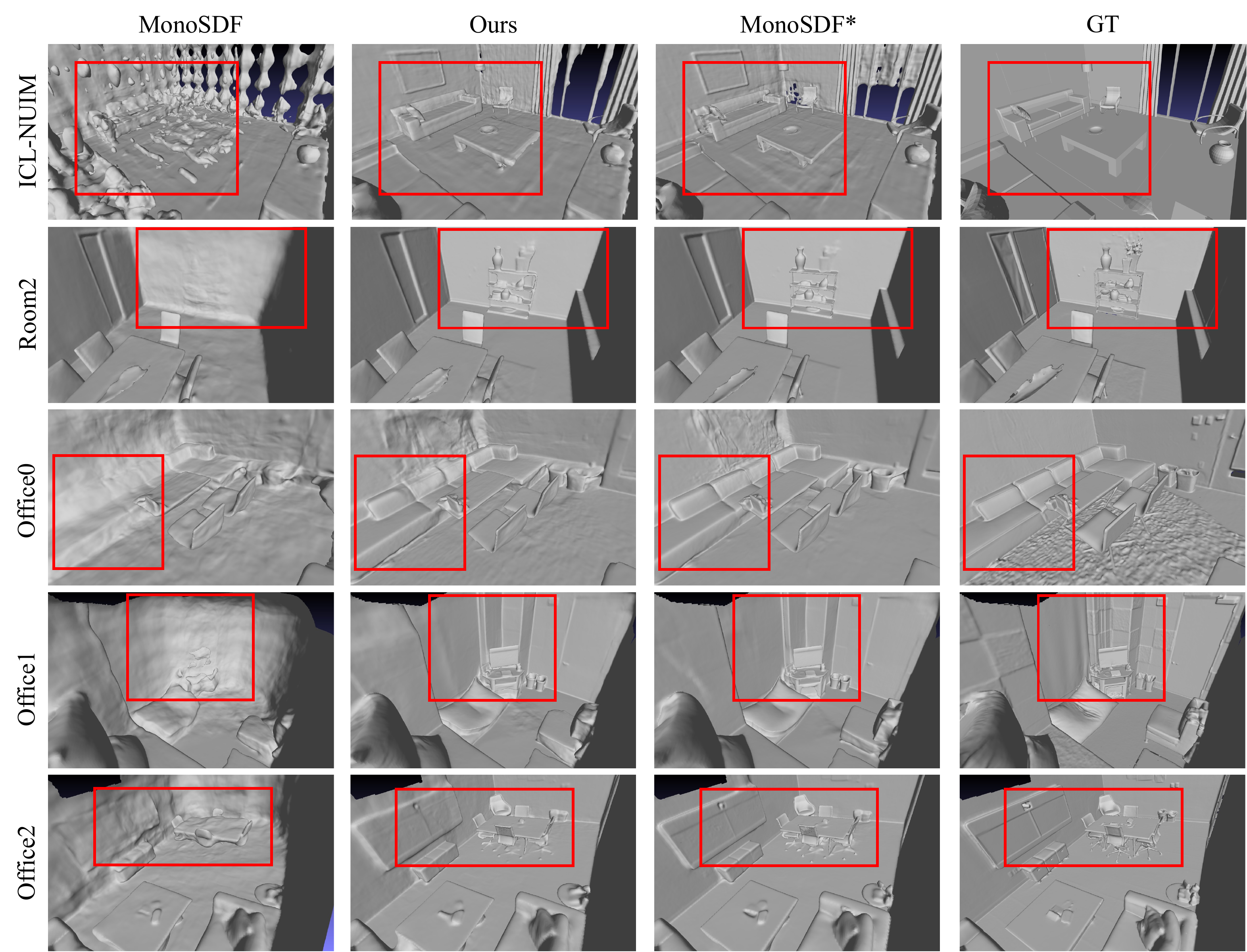}
\vspace{-7mm}
\caption{Qualitative comparison on the ICL-NUIM and Replica datasets. Both `MonoSDF' and `Ours' models are incrementally trained on the 10-step training datasets. The red boxes are the previously learned views. }
\vspace{-4mm}
\label{fig:vis_3d}
\end{figure*}

\vspace{-1mm}
\paragraph{Qualitative Results.}
We further show the qualitative comparison of the ICL-NUIM and Replica datasets in Fig.~\ref{fig:vis_3d}. 
We can see that the MonoSDF baseline learns well for the current scene (scenes outside the red box) but fails on previously seen scenes (highlighted with red boxes) completely. In comparison, our approach is able to reconstruct detailed geometries for both current and previous scenes, achieving similar quality with  ``MonoSDF*'' and ground truth.

\vspace{-1mm}
\subsection{Results on Novel View Synthesis}
We then evaluate our approach for the novel view synthesis task. We adopt NeRF as our backbone and show results for the 360Capture and ScanNet datasets. The four columns $\mathcal{D}^0, \mathcal{D}^3, \mathcal{D}^6, \mathcal{D}^9$ denote the testing dataset at the corresponding time step. The results are obtained from the final model, which has been trained incrementally on all views $\mathcal{D}^{0:9}$. 
Thus $\mathcal{D}^{9}$ is the test dataset of current views and $\mathcal{D}^{0:8}$ are the test datasets of previous views.

\vspace{-1mm}
\paragraph{360Capture.}
We show the results of our approach and baselines on the 360Capture dataset in Tab.~\ref{Tab:exp_real}. 
We can see that the NeRF baseline suffers from the forgetting problem, leading to a large performance drop on the testing data at previous steps $\mathcal{D}^0,\mathcal{D}^3,\mathcal{D}^6$. The incremental baselines (MAS, PackNet, and KR) mitigate the forgetting problem to some extent with limited improvement. 
In comparison, our approach is able to perform consistently well on previous testing data with improvements over the NeRF baseline by 39.6\%, 48.0\%, and 35.9\% for $\mathcal{D}^0,\mathcal{D}^3,\mathcal{D}^6$, and 36.3\% for $\mathcal{D}^{0:9}$ in PSNR.

\vspace{-1mm}
\paragraph{ScanNet.}
We further show results for the more challenging large-scale ScanNet dataset in Tab.~\ref{Tab:exp_scannet}. We can see that both the NeRF baseline and incremental baselines perform poorly on the testing datasets of previous time steps because of the severe forgetting problem caused by little overlap between images in this dataset. Benefiting from the student-teacher pipeline, our method still achieves promising results with significant improvements over incremental baselines (MAS, PackNet, and KR) and SLAM-based baseline iMAP. Comparable performance is also achieved with batch training NeRF*, which further demonstrates the effectiveness of our approach. 

\begin{figure*}[t]
\centering
\vspace{-1mm}
\includegraphics[width=1.0\linewidth]{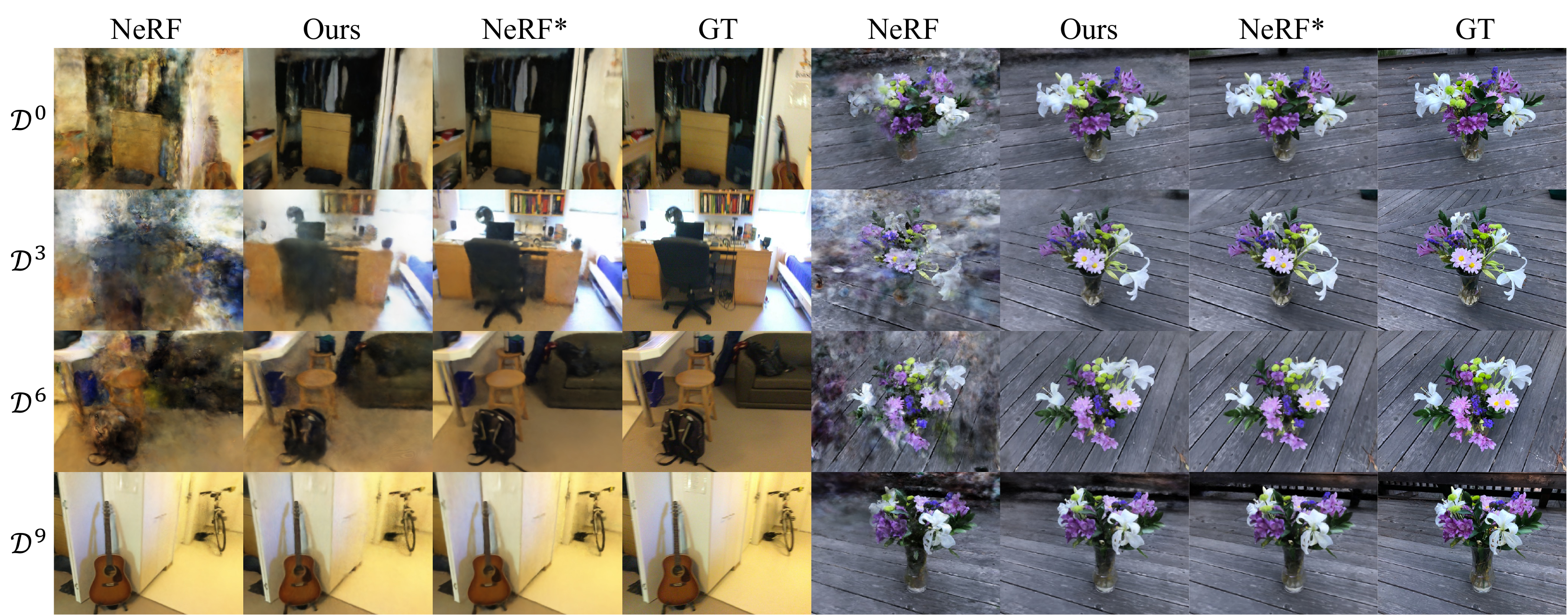}
\vspace{-6mm}
\caption{Qualitative comparison on the ScanNet and 360Capture datasets. `NeRF' and `Ours' models are incrementally trained on the 10-step training datasets. $\mathcal{D}^0,\mathcal{D}^3,\mathcal{D}^6$ denote the results of previous views from each time step test datasets and $\mathcal{D}^9$ is the results of current views from the latest test dataset.
}
\label{fig:vis_real_scan}
\vspace{-2mm}
\end{figure*}

\vspace{-1mm}
\paragraph{Qualitative Results.}
We further show qualitative comparison on the ScanNet scene $101$ and 360Capture scene $Vasdeck$ in Fig.~\ref{fig:vis_real_scan}.
As we can see, the original NeRF suffers from the catastrophic forgetting problem and outputs images on previous time steps $\mathcal{D}^0, \mathcal{D}^3, \mathcal{D}^6$ with severe artifacts including noise and blur. In comparison, our approach generates realistic images with comparable quality to the batch training. This suggests the effectiveness of our proposed approach in mitigating the forgetting problem. 

\begin{table}[t]
\centering
\small
\vspace{-2mm}
\setlength{\tabcolsep}{0.2cm}
\caption{Ablation studies of proposed modules on ICL-NUIM and 360Capture datasets.}
\vspace{-3mm}
\resizebox{0.6\linewidth}{!}{
\begin{tabular}{c|cc|ccc}
\hline
\multirow{2}{*}{Method} & \multicolumn{2}{c|}{ICL-NUIM} & \multicolumn{3}{c}{360Capture} \\
\cline{2-6}
& F1$\uparrow$ & CD$\downarrow$ & PSNR$\uparrow$ & SSIM$\uparrow$ & LPIPS$\downarrow$ \\
\hline
w/o s-t &
${64.71}$ &
${5.94}$ & 
${15.56}$ & 
${0.452}$ & 
${0.405}$  \\
w/o filter &
${81.63}$ &
${3.82}$ &
${19.39}$ & 
${0.639}$ & 
${0.248}$  \\
Ours &
$\textbf{90.32}$ &
$\textbf{2.60}$ & 
$\textbf{21.21}$ & 
$\textbf{0.672}$ & 
$\textbf{0.211}$  \\
\hline
w/ $P^{0:t}$ &
${89.50}$ &
${2.65}$ & 
${21.38}$ & 
${0.673}$ & 
${0.207}$\\
\hline
\end{tabular}
}
\label{Tab:exp_ablation}
\vspace{-2mm}
\end{table}

\begin{table}[t]
\centering
\small
\setlength{\tabcolsep}{0.2cm}
\vspace{-2mm}
\caption{Ablation studies of generalizability to different models on ScanNet dataset.
}
\vspace{-3mm}
\resizebox{0.8\linewidth}{!}{
\begin{tabular}{c|cc}
\hline
{Method} & {Baseline} & {Ours} \\
\hline
NeRF~\cite{mildenhall2020nerf} &
${13.78}$ / ${0.576}$ / ${0.460}$ &
$\textbf{22.59}$ / $\textbf{0.841}$ / $\textbf{0.230}$
\\
Tri-MipRF~\cite{hu2023tri} &
${20.34}$ / ${0.712}$ / ${0.339}$ &
$\textbf{27.02}$ / $\textbf{0.854}$ / $\textbf{0.190}$
\\
ZipNeRF~\cite{barron2023zip} &
${21.78}$ / ${0.739}$ / ${0.302}$ &
$\textbf{27.85}$ / $\textbf{0.876}$ / $\textbf{0.172}$
\\
\hline
\end{tabular}
}
\vspace{-4mm}
\label{Tab:exp_backbone}
\end{table}

\vspace{-1mm}
\subsection{Ablation Study}
\label{sec:abltion}

\vspace{-1mm}
\paragraph{Proposed Module.}
As shown in Tab.~\ref{Tab:exp_ablation}, we conduct ablation studies for both neural surface field and NeRF representations on the ICL-NUIM and 360Capture datasets, respectively. We verify the contribution of our proposed components student-teacher modeling (s-t) and uncertainty-based filter (filter) by removing each component at a time. As can be seen that the performance drops when each component is removed. Specifically, our approach becomes the original NeRF or MonoSDF when the student-teacher modeling is removed, and the model fails completely. Without the uncertainty-based filter, the performance drops significantly for incremental 3D reconstruction task on the ICL-NUIM dataset. This is because the output of the teacher network is not necessarily correct for any input generated from the random inquirer, and the incorrect information can mislead the student during the knowledge distillation. Additionally, we also show results when the camera poses of previous time steps are stored, denoted as w/ $P^{0:t}$. We can see that we achieve comparable performance with this scenario although we do not store any data from the previous time step.  

\vspace{-1mm}
\paragraph{Generalize to different NeRF models.}We also apply our proposed approach to the most recent NeRF models, including Tri-MipRF~\cite{hu2023tri} and ZipNeRF~\cite{barron2023zip}, to show the generalization over different backbones. As shown in Tab.~\ref{Tab:exp_backbone}, we consistently outperform different backbones on ScanNet.

\begin{table}[t]
\centering
\small
\setlength{\tabcolsep}{0.08cm}
\caption{The memory usage analysis for all baselines and Ours on Replica dataset.}
\vspace{-3mm}
\resizebox{\linewidth}{!}{\begin{tabular}{c|cc|cc:ccccccc}
\hline
& MonoSDF~\cite{yu2022monosdf} & MonoSDF*~\cite{yu2022monosdf} & iMAP~\cite{sucar2021imap} & NICE-SLAM~\cite{zhu2022nice} & CNM~\cite{yan2021continual} & MAS~\cite{aljundi2018memory} & PackNet~\cite{mallya2018packnet} & KR~\cite{klein2007parallel} & POD~\cite{douillard2020podnet} & AFC~\cite{kang2022class} & Ours \\
\hline
Memory(MB) & ${0}$ & ${300}$ & ${30}$ & ${30}$ & $\textbf{2}$ & ${16}$ & ${11}$ & ${30}$ & ${33}$ & ${19}$ & $\underline{3}$ \\
\hline
\end{tabular}}
\label{Tab:memory}
\vspace{-4mm}
\end{table}




\begin{figure}[t]
\centering
\vspace{-1mm}
\includegraphics[width=.75\linewidth]{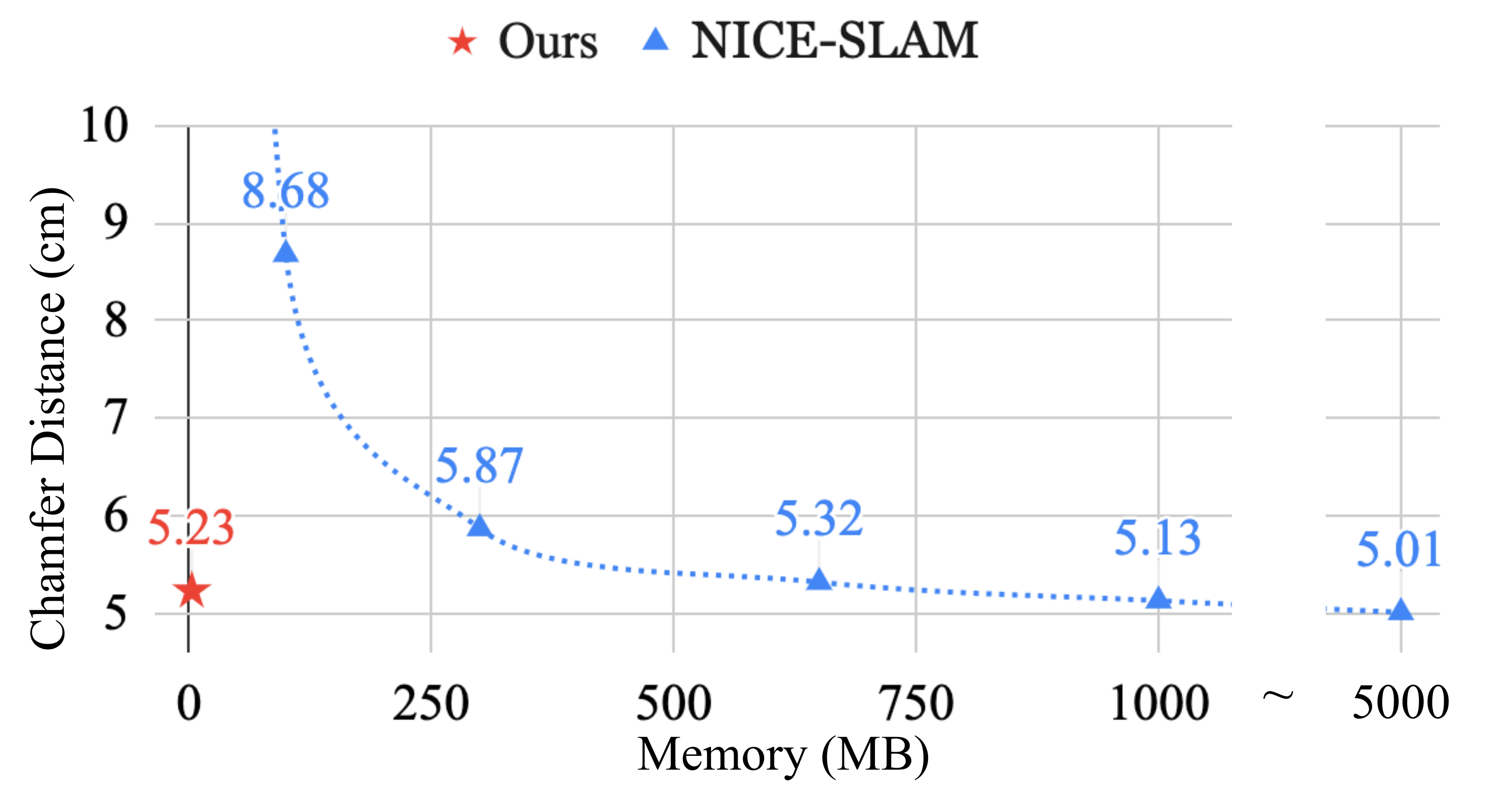}
\vspace{-5mm}
\caption{Comparison of Ours and NICE-SLAM with different memory usage.} 
\vspace{-4mm}
\label{fig:memory}
\end{figure}

\vspace{-1mm}
\subsection{Memory Analysis}
\label{sec:memory}
As Tab.~\ref{Tab:memory} shown, we further provide the memory usage analysis for all methods presented in Tab.~\ref{Tab:exp_ICL} and Tab.~\ref{Tab:exp_replica}. Among the baselines, iMAP, NICE-SLAM, CNM, KR, and POD require additional space to store keyframes, MAS, PackNet, and AFC need memory to save the importance score or masks for network parameters. POD, AFC, and ours use a distillation strategy that stores the teacher model. The batch training method MonoSDF* requires much more memory than other baselines for storing a full batch of data. We can conclude that our model achieves better performance with smaller memory usage based on the memory usage (Tab.~\ref{Tab:memory}) and the performance comparison (Tab.~\ref{Tab:exp_replica}).

Memory replay technique plays a crucial role in mitigating the forgetting problem in existing SLAM-based methods~\cite{sucar2021imap, zhu2022nice} and incremental NIRs~\cite{yan2021continual, iccv23clnerf}. These approaches require additional memory allocation to store the previously seen data. 
To understand the relationship between memory utilization and performance, we conducted experiments using the strongest baselines NICE-SLAM with varying memory capacities on the large-scale scene Apartment~\cite{zhu2022nice}, as shown in Fig.~\ref{fig:memory}. We can see that performance improves as memory usage increases from 0 to 5,000MB (equivalent to the batch training method MonoSDF*). In comparison, our approach performs similarly to MonoSDF* (5.12cm) while maintaining minimal memory consumption (around 3MB).

\vspace{-1mm}
\section{Conclusion}
\vspace{-1mm}
In this paper, we explore the task of incremental learning for Neural Implicit Representations (NIRs). We propose a student-teacher pipeline for mitigating the catastrophic forgetting problem. To improve the effectiveness of the data provided by the teacher network, we further design a random inquirer and an uncertainty-based filter for useful knowledge distillation. 
Supervised learning and knowledge distillation are iteratively utilized for the combination of preserving old information and learning current new data. 
Experiments on both 3D reconstruction and novel view synthesis demonstrate that our model achieves great improvement compared to baselines under the incremental setting.

\paragraph{Acknowledgement.} This research work is supported by the Agency for Science, Technology and Research (A*STAR) under its MTC Programmatic Funds (Grant No. M23L7b0021).

%
%

\bibliographystyle{IEEEtran}

\clearpage


\appendix
\renewcommand\thefigure{S\arabic{figure}}
\renewcommand\thetable{S\arabic{table}}

\section{Implementation Details}

\subsection{Method Details}
\paragraph{NeRF Sampling.}
We sample points along the ray represented as $\textbf{r}(t) = \textbf{o} + t\textbf{d}$, where $\textbf{o}$ is the camera center and $\textbf{d}$ is the ray direction. For each dataset, the hyperparameters 'near' $t_n$ and 'far' $t_f$ specify the sampling interval of the ray. The sampled points along the ray are thus constrained within this interval. Consequently, the distance represents as $\delta_{i} = t_{i+1} - t_{i}$. 

\paragraph{Loss Function.}
We provide more details about the $\mathcal{L}_{eik}$ and $\mathcal{L}_{prior}$ losses in Eqn.~(2) in the main paper.
\begin{equation}
\begin{aligned}
\mathcal{L}_{eik} = \sum (\nabla f_{\theta}(x))^2,\quad \mathcal{L}_{prior} = \mathcal{L}_{d} + \mathcal{L}_{n}.
\end{aligned}
\end{equation}
where $\mathcal{L}_{eik}$ is the SDF regularizer for neural surface field~\cite{wang2021neus, yariv2021volume, yu2022monosdf},  $x$ denotes the sampled points, and $\mathcal{L}_{prior}$ represents the optional geometry prior term~\cite{yu2022monosdf}. Specifically, the depth consistency loss $\mathcal{L}_{d}$ takes the form of an L2 loss involving shift $q$ and scale $w$: \begin{equation}
\begin{aligned}
\mathcal{L}_{d} = \sum ||w\hat{D}(r)+q-\bar{D}(r)||^2,
\\
\hat{D}(r) = \sum_{i=1}^{N}T_{i}\big(1-\exp(-\sigma_{i}\delta_{i})\big)t_{i}.
\end{aligned}
\end{equation}
We solve for $w$ and $q$ with a least-squares criterion which has a closed-form solution.
The normal consistency loss $\mathcal{L}_{n}$ consists of angular and L1 losses:
\begin{equation}
\begin{aligned}
\mathcal{L}_{n} = \sum ||\hat{N}(r)-\bar{N}(r)||_{1} + ||1-\hat{N}(r)^T\bar{N}(r)||_{1},
\\
\hat{N}(r) = \sum_{i=1}^{N}T_{i}\big(1-\exp(-\sigma_{i}\delta_{i})\big)\hat{n}_{i}.
\end{aligned}
\end{equation}

\section{Additional Experiments}

\subsection{Dataset Details}
We conduct experiments on four datasets, NeRF-real360~\cite{mildenhall2020nerf}, ScanNet~\cite{dai2017scannet}, Replica~\cite{straub2019replica} and ICL-NUIM~\cite{handa2014ICL}. 
They contain sequences of RGB(D) images extracted from a video with camera poses. To simulate the incremental scenarios, we divide the dataset images and camera poses into ten groups according to the time sequence of the video and then randomly sample some frames in each group as the test set. To use MonoSDF~\cite{yu2022monosdf}, we follow their setting and use the pre-trained Omnidata model~\cite{eftekhar2021omnidata} to predict the monocular depth and normal map as part of the dataset.

\paragraph{NeRF-real360.} We use both two scenes with around 100 images with camera poses in each scene. Following the setting in NeRF~\cite{mildenhall2020nerf}, we randomly sample 12.5\% of images as the test set and use the rest for training.
For incremental setting, we split one scene into 10-time steps $\mathcal{D}=\{\mathcal{D}^0,\mathcal{D}^1,\cdots,\mathcal{D}^9\}$ in the temporal order. 

\paragraph{ScanNet.} Following \cite{liu2020neural}, we use scene $101$ and scene $241$ with $>1000$ images at a resolution of $648\times484$. We randomly sample 20\% of images as the training set and use the rest for testing. The camera poses are optimized by COLMAP \cite{schoenberger2016mvs, schoenberger2016sfm}. For incremental setting, we split a scene into 10-time steps $\mathcal{D}=\{\mathcal{D}^0,\mathcal{D}^1,\cdots,\mathcal{D}^9\}$ in the temporal order.

\paragraph{Replica.} 
Following \cite{yu2022monosdf}, we use eight scenes from the Replica dataset with 100 images with camera poses in each scene. For incremental setting, we split one scene into 10-time steps $\mathcal{D}=\{\mathcal{D}^0,\mathcal{D}^1,\cdots,\mathcal{D}^9\}$ in the temporal order. 

\paragraph{ICL-NUIM.} 
We use about 200 images with camera poses and depths. For incremental setting, we split one scene into 10-time steps $\mathcal{D}=\{\mathcal{D}^0,\mathcal{D}^1,\cdots,\mathcal{D}^9\}$ in the temporal order. 

\subsection{Quantitative Results}

\paragraph{3D reconstruction results on ScanNet.}
As shown in Tab.~\ref{Tab:exp_3D_scannet}, we compare with MonoSDF, upper bound method MonoSDF*, the best two baselines NICE-SLAM and KR in Tab.~2 of the main paper. We can see that our approach improves over MonoSDF by 86.0\% for F1 and outperforms NICE-SLAM and KR significantly. Moreover, we achieve comparable performance with batch-training MonoSDF* although we do not use any previous data.

\begin{table}
\centering
\small
\caption{Comparison with baselines on the ScanNet. (Best and second best results are highlighted in bold and underlined, respectively.)}
\setlength{\tabcolsep}{0.08cm}
\resizebox{0.9\linewidth}{!}{\begin{tabular}{c|cc|c:cc}
\hline
& MonoSDF~\cite{yu2022monosdf} & MonoSDF*~\cite{yu2022monosdf} & NICE-SLAM~\cite{zhu2022nice} & KR~\cite{klein2007parallel} & Ours \\
\hline
F1$\uparrow$ & ${39.48}$ & ${73.32}$ & $\underline{67.04}$ & ${59.88}$ & $\textbf{73.35}$ \\
CD$\downarrow$ & ${12.3}$ & ${4.21}$ & $\underline{5.09}$ & ${6.93}$ & $\textbf{4.15}$ \\
\hline
\end{tabular}}
\label{Tab:exp_3D_scannet}
\end{table}

\paragraph{More results on Replica.}
We have included the average quantitative results on the Replica dataset in Tab.~2 of the main paper. For a comprehensive analysis, we further provide the detailed quantitative results for all scenes in Tab.~\ref{Tab:sup_exp_replica}. It is evident from the results that our approach outperforms MonoSDF by a significant margin across all scenes and achieves comparable performance to the upper bound batch training method `MonoSDF*'.

\begin{table*}[htb]
\centering
\small
\setlength{\tabcolsep}{0.08cm}
\caption{{Comparison with baselines on the replica dataset. (Best and second best results are highlighted in bold and underlined, respectively.)
}}
\vspace{-1mm}
\resizebox{\linewidth}{!}{\begin{tabular}{c|cc|cc:ccccccc}
\hline
CD$\downarrow$ & MonoSDF~\cite{yu2022monosdf} & MonoSDF*~\cite{yu2022monosdf} & iMAP~\cite{sucar2021imap} & NICE-SLAM~\cite{zhu2022nice} & CNM~\cite{yan2021continual} & MAS~\cite{aljundi2018memory} & PackNet~\cite{mallya2018packnet} & KR~\cite{klein2007parallel} & POD~\cite{douillard2020podnet} & AFC~\cite{kang2022class} & Ours \\
\hline
room0 & ${3.94}$ & ${2.45}$ & ${4.32}$ & $\underline{2.80}$ & ${4.49}$ & ${3.78}$ & ${4.59}$ & ${2.83}$ & ${3.42}$ & ${3.10}$ & $\textbf{2.49}$ \\
room1 & ${6.24}$ & ${2.97}$ & ${4.28}$ & $\underline{2.53}$ & ${5.28}$ & ${6.24}$ & ${5.72}$ & $\textbf{2.49}$ & ${3.09}$ & ${2.73}$ & ${2.54}$ \\
room2 & ${7.74}$ & ${2.99}$ & ${5.10}$ & $\textbf{2.83}$ & ${6.92}$ & ${6.72}$ & ${7.29}$ & ${3.48}$ & ${4.06}$ & ${3.76}$ & $\underline{2.88}$ \\
office0 & ${7.96}$ & ${3.33}$ & ${5.99}$ & $\textbf{2.14}$ & ${7.37}$ & ${7.44}$ & ${6.86}$ & $\underline{4.11}$ & ${4.82}$ & ${4.69}$ & ${4.32}$ \\
office1 & ${11.63}$ & ${2.77}$ & ${4.49}$ & $\textbf{2.43}$ & ${6.90}$ & ${12.78}$ & ${8.69}$ & ${4.13}$ & ${4.61}$ & ${4.29}$ & $\underline{2.85}$ \\
office2 & ${12.93}$ & ${3.18}$ & ${5.19}$ & $\underline{3.70}$ & ${8.21}$ & ${11.55}$ & ${11.42}$ & ${6.66}$ & ${6.93}$ & ${6.24}$ & $\textbf{3.51}$ \\
office3 & ${11.09}$ & ${3.22}$ & ${4.86}$ & $\underline{3.67}$ & ${6.12}$ & ${9.84}$ & ${7.90}$ & ${4.37}$ & ${5.14}$ & ${4.72}$ & $\textbf{3.48}$ \\
office4 & ${7.12}$ & ${2.64}$ & ${5.71}$ & $\underline{3.31}$ & ${7.03}$ & ${5.47}$ & ${7.44}$ & ${3.88}$ & ${4.39}$ & ${4.03}$ & $\textbf{2.81}$ \\
\hline
Average & ${8.58}$ & ${2.94}$ & ${4.99}$ & $\textbf{2.93}$ & ${6.54}$ & ${7.98}$ & ${7.49}$ & ${3.99}$ & ${4.56}$ & ${4.20}$ & $\underline{3.11}$ \\
\hline
\end{tabular}}
\label{Tab:sup_exp_replica}
\vspace{-2mm}
\end{table*}

\begin{figure*}
\centering
\includegraphics[width=.95\linewidth]{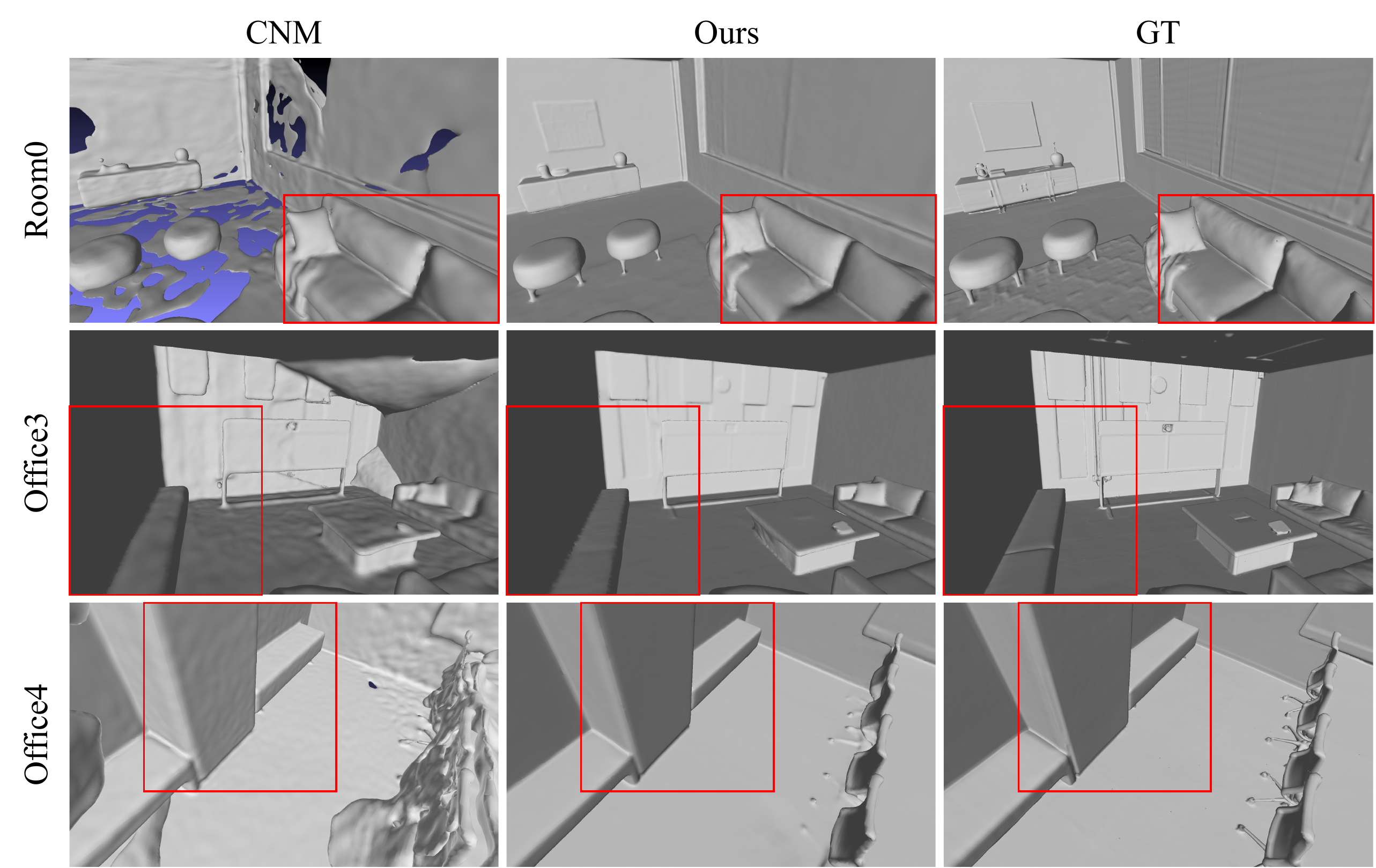}
\caption{Visual comparison between CNM and our approach on the Replica dataset. Both `CNM' and  `Ours' models are incrementally trained using the 10-step training datasets. The red boxes denote current views, and the rest are the previous views.}
\label{fig:vis_cnm}
\end{figure*}

\begin{table}[t]
\centering
\small
\setlength{\tabcolsep}{0.2cm}
\caption{\textcolor{black}{Ablation study of the proposed iterative optimization mechanism on ICL-NUIM.
}}
\vspace{-3mm}
\resizebox{0.9\linewidth}{!}{\begin{tabular}{c|ccccccc}
\hline
Sup. Iter. : Dis. Iter. & 9:1 & 4:1 & 2:1 & 1:1 & 1:2 & 1:4 & 1:9 \\
\hline
F1$\uparrow$ & 
${84.62}$ & 
${86.77}$ & 
${88.57}$ & 
$\textbf{90.32}$ & 
${89.01}$ & 
${88.47}$ & 
${86.85}$  \\
CD$\downarrow$ &
${3.23}$ &
${2.95}$ & 
${2.75}$ & 
$\textbf{2.60}$ & 
${2.78}$ & 
${2.86}$ & 
${2.90}$  \\
\hline
\end{tabular}}
\label{Tab:exp_loss}
\vspace{-1mm}
\end{table}

\subsection{Ablation Studies.}
\paragraph{Proposed Optimization Mechanism.}
As shown in Tab.~\ref{Tab:exp_loss}, we conduct the ablation study of the iterative optimization mechanism (Eqn.~(7))) on the ICL-NUIM dataset, where $x:y$ denotes $x$ iteration(s) of the supervised training followed by $y$ iteration(s) of distillation training. Our chosen strategy (1:1) gives optimal results, which effectively maintains equilibrium between learning new batch data and safeguarding against the forgetting of previously learned data. Conversely, (9:1) and (1:9) result in overemphasis on either the new data or the previously learned data, thus leading to detrimental impacts on the overall outcomes.

\subsection{Comparison to CNM.}
The most related work to ours is Continual Neural Mapping (CNM). We reimplemented CNM for comparison with our method since they did not open-source their code. It is important to highlight that CNM primarily focuses on SDF optimization which lacks a rendering model, thus making it less suitable for novel view synthesis tasks. Consequently, our evaluation is limited to the 3D reconstruction datasets ICL-NUIM and Replica.

We show the results of the evaluation in Tab.~1 and Tab.~2 of the main paper. We can see the superior performance of our model in comparison with CNM. Specifically, we achieved a significant F1 score advantage of 19 and 20.39 on Replica and ICL-NUIM, respectively. 
Visual comparisons are also depicted in Fig.~\ref{fig:vis_cnm}. It is evident from these visualizations that CNM encounters challenges in addressing holes and exhibits perceptible blurriness in the reconstructed scenes due to the limitations of their fixed-size replay buffer, which fails to adequately encompass the full distribution of previous data. In contrast, our method leverages the pipeline of a student-teacher model integrated with a random inquirer, thereby ensuring the preservation of the capability to query all observations from previous data.

\subsection{Comparison to the upper bound method.}
\begin{figure}[t]
\centering
\includegraphics[width=0.98\linewidth]{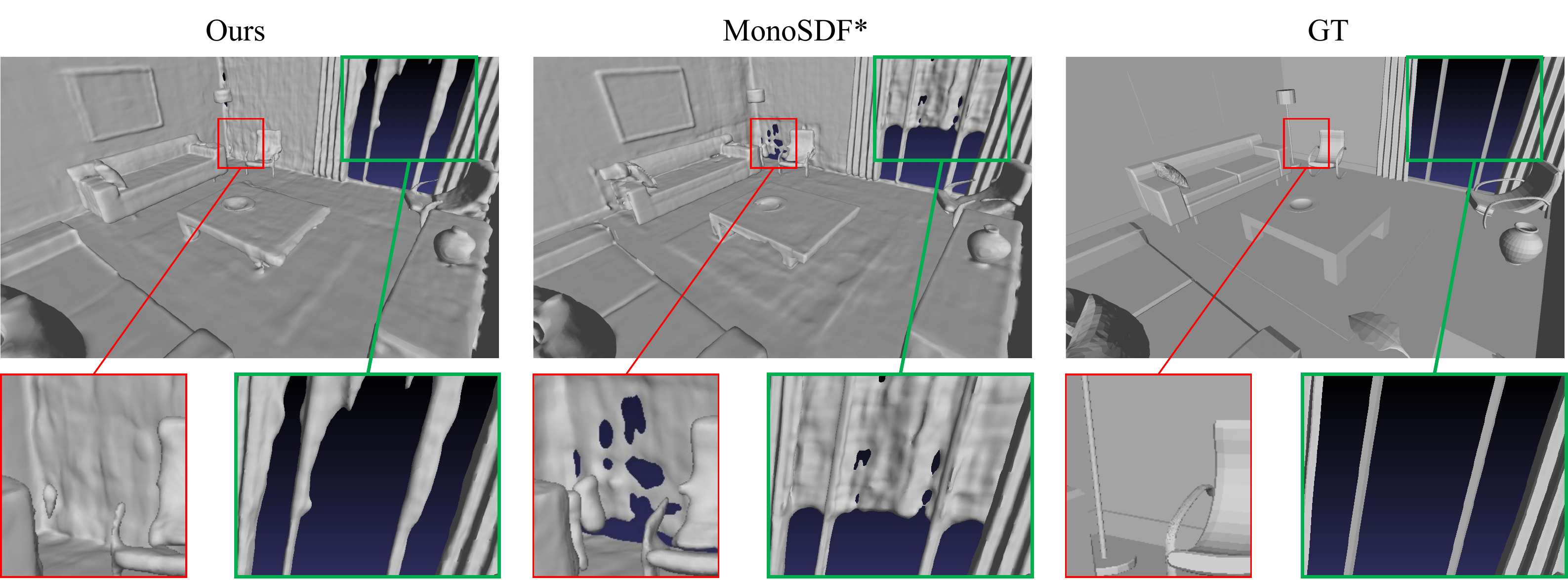}
\caption{Comparison between `Ours' and the upper bound method `MonoSDF*' on ICL-NUIM.}
\label{fig:compare_up}
\end{figure}

We show a more detailed comparison between our method and the batch-training method MonoSDF* (upper bound method) in Fig.~\ref{fig:compare_up}. We can see that MonoSDF* produces holes on the surface and also artifacts on the window and wall, while our method predicts complete surface to the ground truth.

\begin{figure}[h]
\centering
\includegraphics[width=0.98\linewidth]{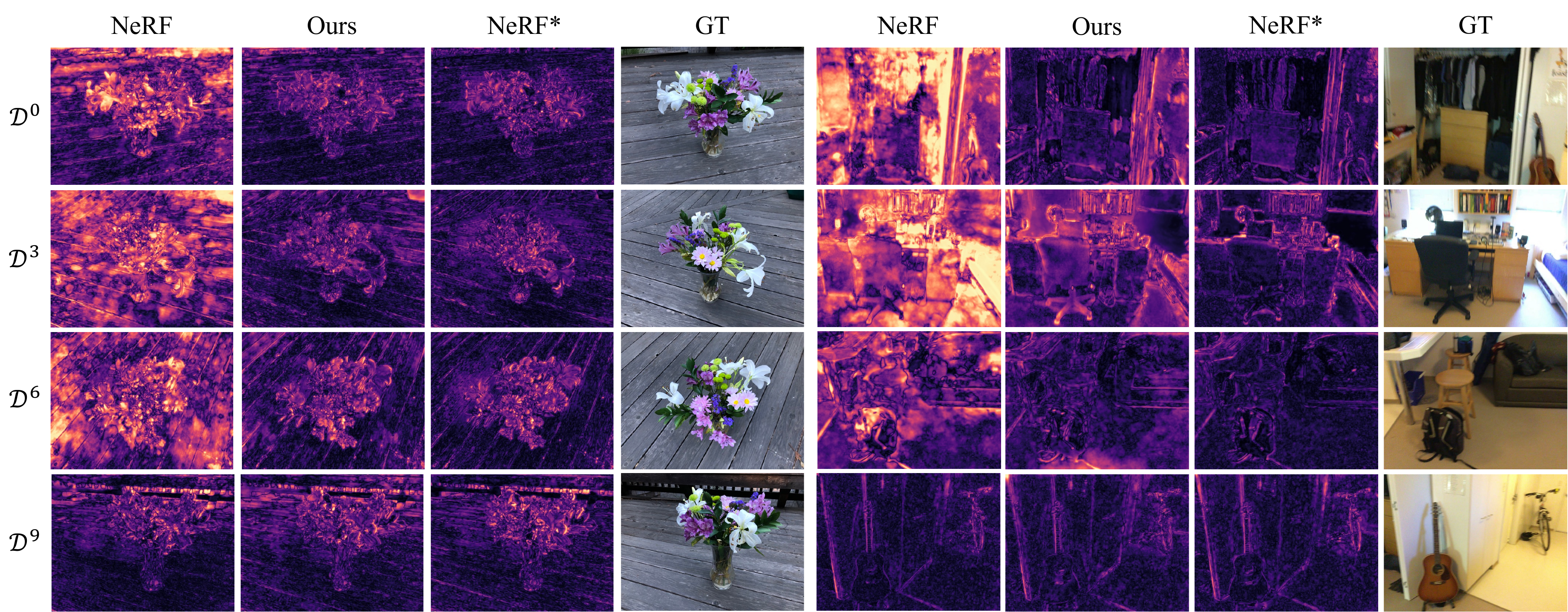}
\caption{The visualization of the error area of `Ours' and the upper bound method `NeRF*'.}
\label{fig:flip_img}
\end{figure} 

\subsection{More Qualitative Results}

\paragraph{Error Area Visualization}
We show the visualization of the error area using FLIP~\cite{Andersson2021b} in Fig.~\ref{fig:flip_img}. The baseline NeRF suffers from the catastrophic forgetting problem, evident from the large error areas in the results of $\mathcal{D}^{0,3,6}$. In comparison, our method has much smaller error areas in $\mathcal{D}^{0,3,6}$ and achieves comparable performance with the batch-training method NeRF* (the upper bound method).

\begin{figure*}
\centering
\includegraphics[width=.95\linewidth]{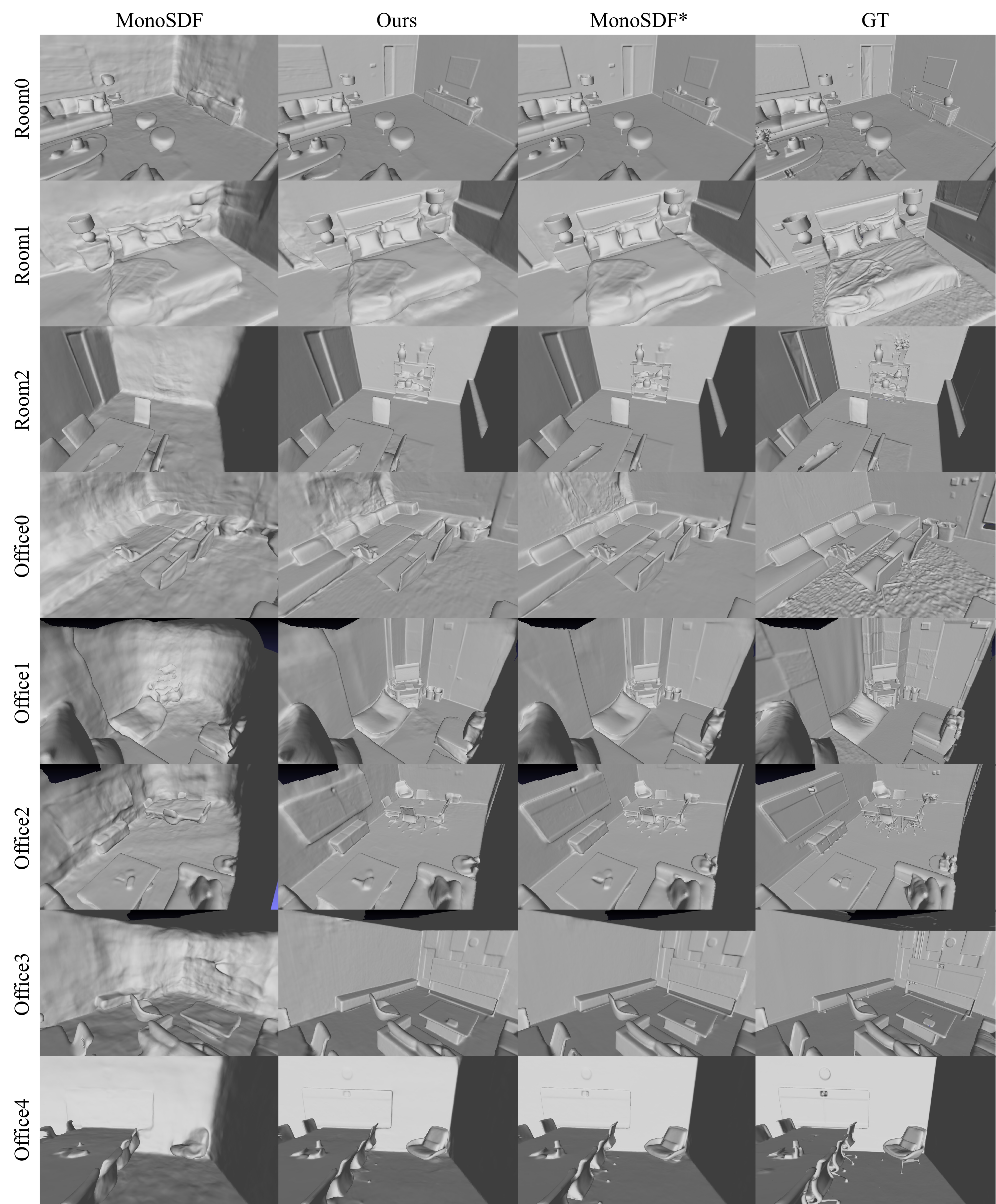}
\caption{Qualitative comparison with the baseline `MonoSDF' and upper bound batch training method `MonoSDF*' on eight scenes of the Replica dataset. `GT' represents the ground truth images. `MonoSDF' and `Ours' models are incrementally trained on the 10-step training datasets.}
\label{fig:vis_sup_replica}
\end{figure*}

\begin{figure*}
\centering
\includegraphics[width=.95\linewidth]{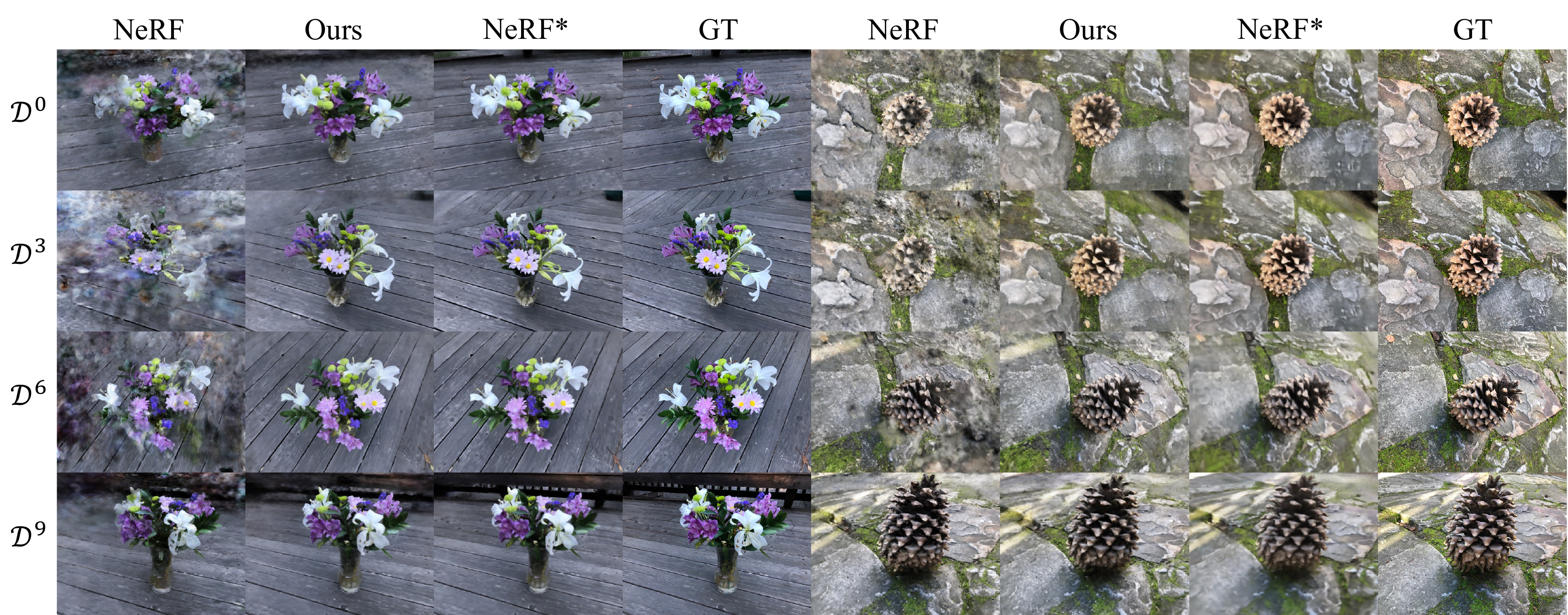}
\caption{Qualitative comparison with the baseline `NeRF' and upper bound batch training method `NeRF*' on scene $Vasdeck$ (left) and $Pinecone$ (right) of NeRF-real360 dataset. `GT' represents the ground truth images and $\mathcal{D}^0,\mathcal{D}^3,\mathcal{D}^6,\mathcal{D}^9$ denote the results from each time step test views. `NeRF' and `Ours' models are incrementally trained on the 10-step training datasets.}
\label{fig:vis_sup_real}
\end{figure*}

\paragraph{More Visualization}
The results presented in Fig.~\ref{fig:vis_sup_replica} showcase the 3D reconstruction performances of the baseline `MonoSDF', our proposed approach, and the upper bound batch training method `MonoSDF*'. Similarly, Fig.~\ref{fig:vis_sup_real} and Fig.~\ref{fig:vis_sup_scan} demonstrate the novel view synthesis performances of the baseline `NeRF', our proposed approach, and the upper bound batch training method `NeRF*'. 
Through these visualizations, we can conclude that our method consistently addresses the catastrophic forgetting problem and achieves comparable results to the upper bound batch training methods in both the novel view synthesis and 3D reconstruction scenarios.

\begin{figure*}
\centering
\includegraphics[width=.95\linewidth]{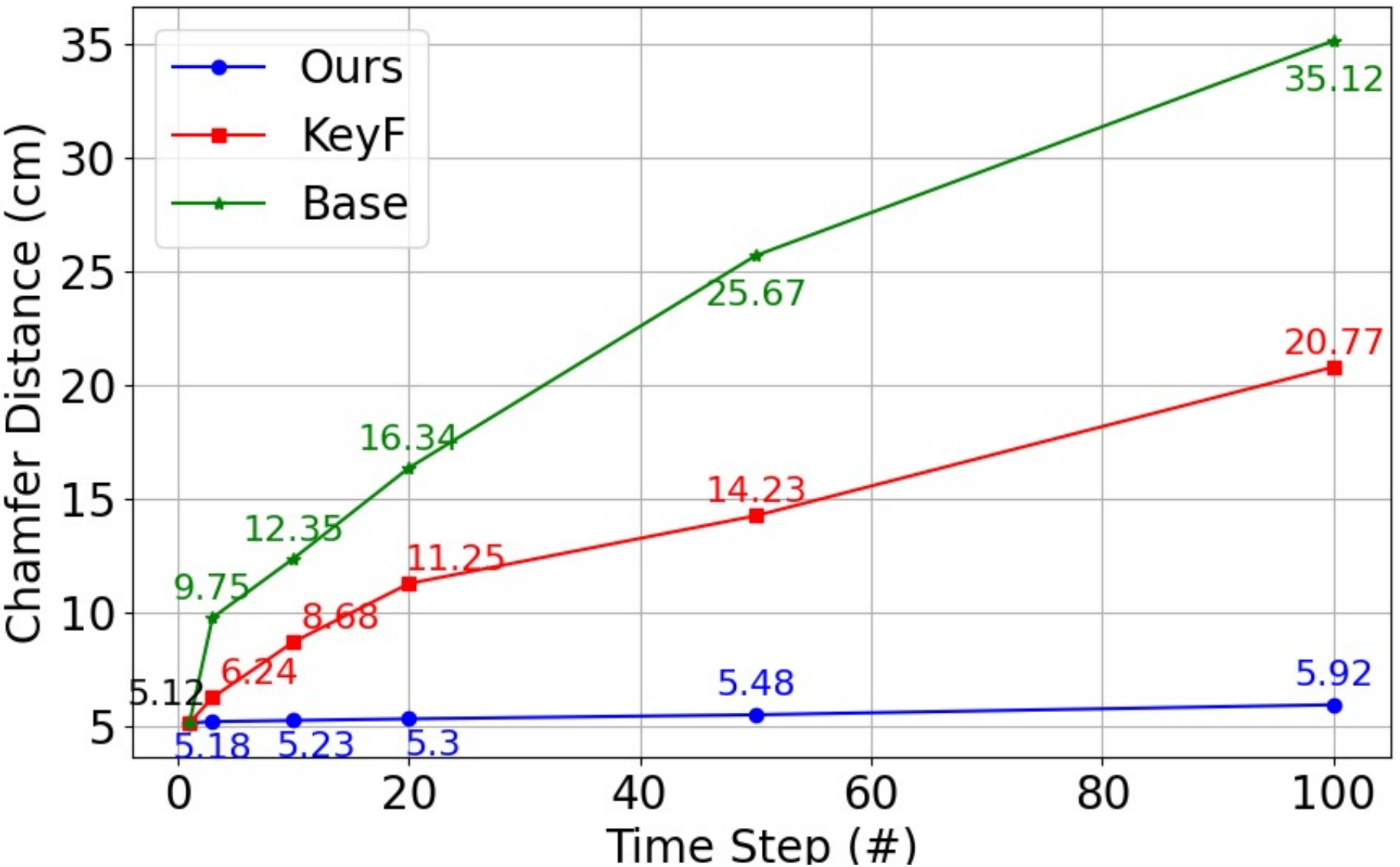}
\caption{Time step comparison.}
\label{fig:vis_steps}
\end{figure*}

\subsection{More Visualization}
Training large-scale scenes with 24/7 streaming data from robots or devices requires many, possibly infinite, time steps. In such scenarios, methods that store even one keyframe per step demand enormous data storage. We set our evaluation protocol to 10 time steps to validate our method.  
As shown in Fig.~\ref{fig:vis_steps}, we further conduct experiments with more time steps. Our method's reconstruction error remains almost constant as the number of time steps increases, while KeyF's error increases sharply due to increasingly sparse keyframes.

\begin{figure*}
\centering
\includegraphics[width=.95\linewidth]{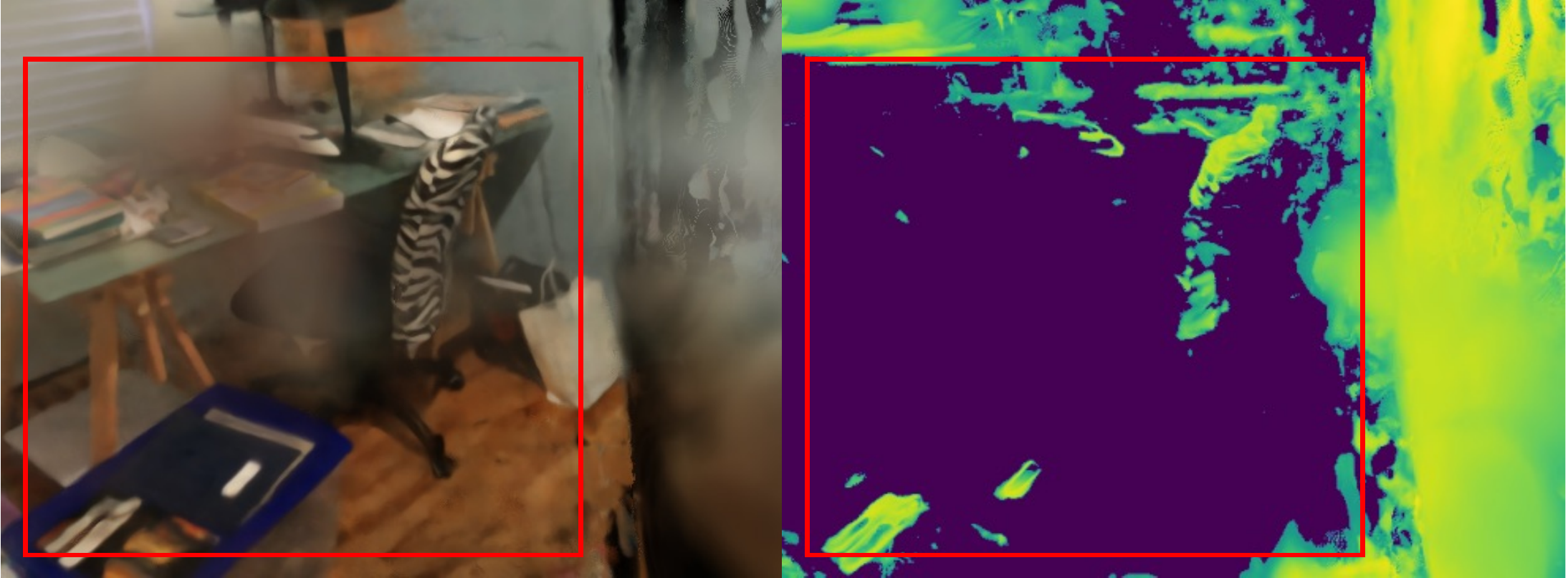}
\caption{Visualization of RGB Image with Uncertainty Map (Normalized).}
\label{fig:vis_uncert}
\end{figure*}

\begin{figure*}
\centering
\includegraphics[width=.95\linewidth]{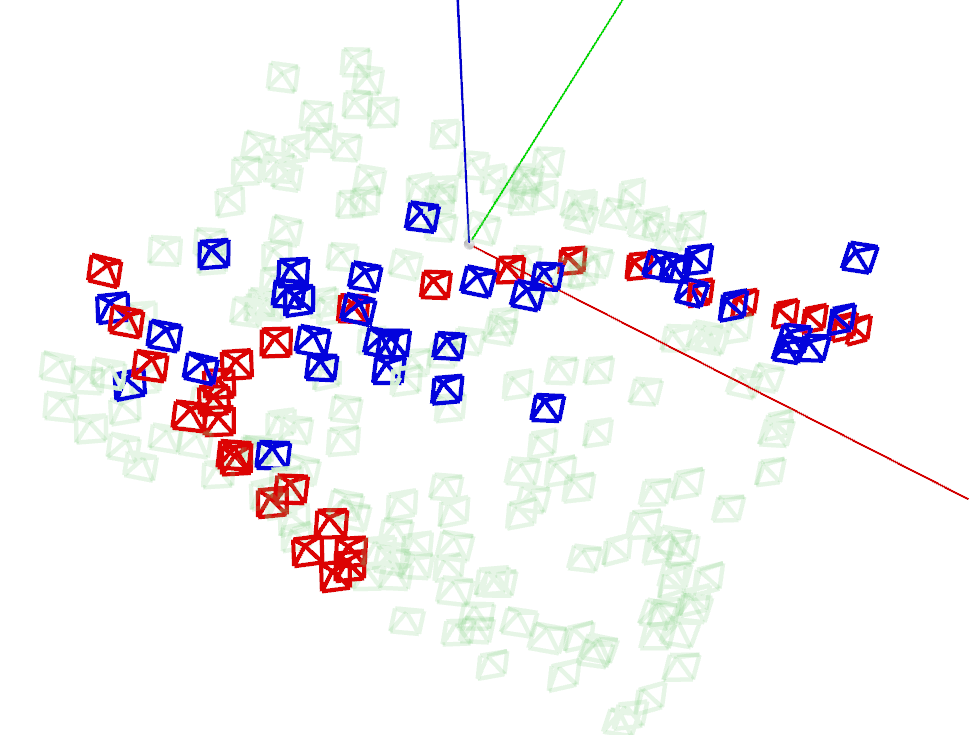}
\caption{Uncertainty-based Pose Filtering.}
\label{fig:vis_pose}
\end{figure*}

\subsection{Discussion about Uncertainty Map.}
The uncertainty score filters camera views to select useful knowledge from the teacher model (\cf the main paper L246). To verify this, we show a rendered RGB image with an uncertainty map in Fig.~\ref{fig:vis_uncert} Previously seen areas (in red box) with lower uncertainty (in purple) have better RGB quality. 
As shown in Fig.~\ref{fig:vis_pose}, the filtered samples (in blue) are accurately selected from randomly generated poses (in green) using our uncertainty filtering method. These filtered samples effectively cover the target poses (in red) area and the camera directions are generally consistent. This demonstrates the effectiveness of our uncertainty filter.

\begin{figure*}
\centering
\includegraphics[width=.95\linewidth]{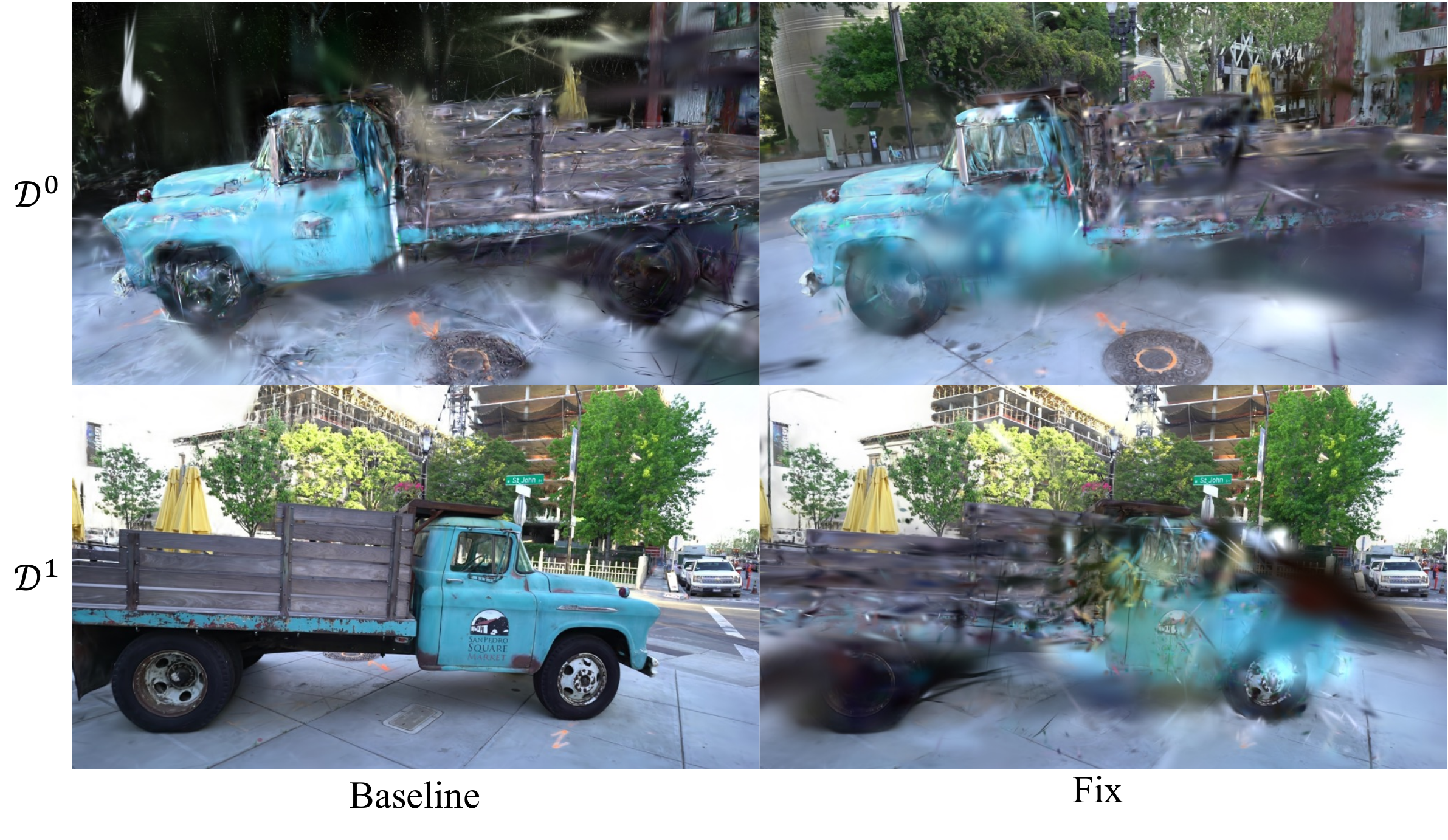}
\caption{Incremental 3DGS. $\mathcal{D}^0, \mathcal{D}^1$ are previous learned and current views.}
\label{fig:vis_gs}
\end{figure*}

\subsection{Discussion about explicit representations (3DGS)}
Experiments with incremental 3DGS (Fig.~\ref{fig:vis_gs}) show that the baseline suffers from catastrophic forgetting in learned views ($\mathcal{D}^0$), and the method of fixing learned points causes severe conflicts in overlapping areas (\eg, the foreground truck in both $\mathcal{D}^0$ and $\mathcal{D}^1$) 
This conflict arises because the fixed points from previously learned data already occupy the spatial positions where new points from future observations should be placed. These fixed points have colors that are correct for the previous views but incorrect for the new views. Consequently, learning new views becomes challenging due to this spatial and color conflict between existing and new points. Newly generated Gaussian points from new observations are unable to properly represent the scene as their optimal positions are already occupied by fixed points with potentially incompatible colors. Splitting certain points without conflict is an interesting problem worth exploring to address this issue.

Moreover, recent works like Scaffold-GS~\cite{lu2024scaffold} and VastGaussian~\cite{lin2024vastgaussian} introduce implicit networks (MLP) for better quality. The same forgetting problems arise with implicit neural fields, making our method a viable solution.

\section{Rendered Video}
We offer rendered videos `\textbf{ScanNet101.mp4}' for the novel view synthesis task on the ScanNet dataset and video `\textbf{ICL\_NUIM.mp4}' for the 3D reconstruction task on the ICL-NUIM dataset. In these videos, the terms 'NeRF' and 'MonoSDF' refer to the results obtained from the baseline model, while 'Ours' represents the outcomes generated by our proposed model. These videos showcase rendered views from $\mathcal{D}^0$ to $\mathcal{D}^9$. 

Upon comparing these three videos, it becomes evident that the baseline models (NeRF and MonoSDF) suffer from the catastrophic forgetting problem, leading to low-quality videos with noticeable artifacts such as noise and blur. In contrast, our approach significantly mitigates this problem through the utilization of our proposed teacher-student pipeline, resulting in videos of higher quality and reduced artifacts.

\section{Limitation}
\begin{figure*}
\centering
\includegraphics[width=.95\linewidth]{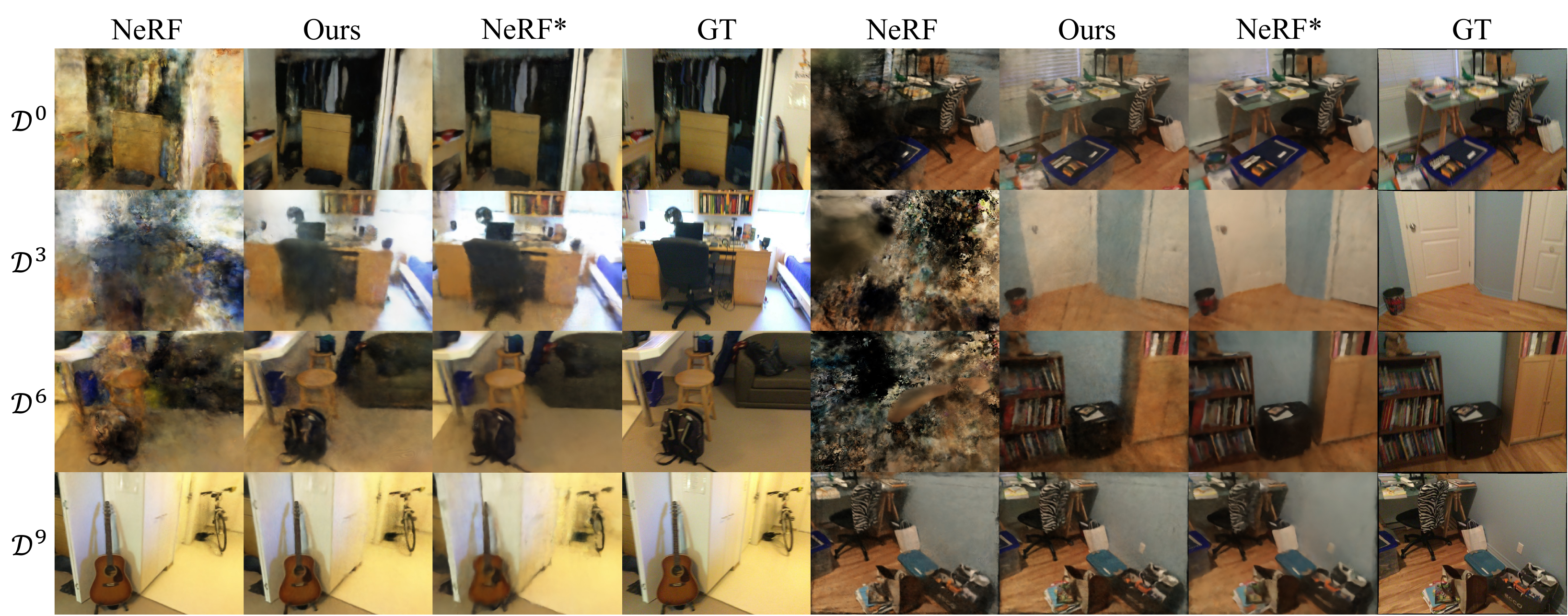}
\caption{Qualitative comparison with the baseline `NeRF' and upper bound batch training method `NeRF*' on scene $101$ (left) and $241$ (right) of ScanNet dataset. `GT' represents the ground truth images and $\mathcal{D}^0,\mathcal{D}^3,\mathcal{D}^6,\mathcal{D}^9$ denote the results from each time step test views. `NeRF' and `Ours' models are incrementally trained on the 10-step training datasets.}
\label{fig:vis_sup_scan}
\end{figure*}

While our method demonstrates a substantial improvement over the baselines in the context of incremental training, there is still room for further enhancement compared to models trained in batch mode. The large-scale nature of the scene provides ample information, and the camera facing-outwards dataset exhibits minimal overlap between views. Consequently, our method may not be able to memorize all the knowledge from the limited images. Even the model trained with batch training can only acquire a limited amount of information about the scene, as illustrated in Fig.~\ref{fig:vis_sup_scan}. Therefore, there is potential for further progress in capturing a more comprehensive understanding of the scene.

\end{document}